\documentclass{article}

\usepackage{microtype}
\usepackage{graphicx}
\usepackage{subcaption}
\usepackage{booktabs} 

\usepackage{hyperref}

\usepackage[preprint]{icml2026}

\usepackage{amsmath}
\usepackage{amssymb}
\usepackage{mathtools}
\usepackage{amsthm}

\usepackage[capitalize,noabbrev]{cleveref}

\usepackage{graphicx}
\usepackage{xcolor}
\usepackage{notation}
\usepackage{researchpack}
\usepackage{hyperref}
\usepackage[capitalize,noabbrev]{cleveref}
\usepackage{booktabs}
\usepackage{subcaption}
\usepackage{natbib}
\usepackage{multirow}

\theoremstyle{plain}
\newtheorem{theorem}{Theorem}[section]
\newtheorem{proposition}[theorem]{Proposition}
\newtheorem{lemma}[theorem]{Lemma}

\theoremstyle{definition}
\newtheorem{definition}[theorem]{Definition}
\newtheorem{remark}[theorem]{Remark}

\definecolor{posgreen}{HTML}{1C7935}
\definecolor{lightgreen}{HTML}{D9EFD3}
\definecolor{baselinegray}{HTML}{F0F0F0}
\definecolor{negred}{HTML}{B42318}
\usepackage[table]{xcolor}

\usepackage[textsize=tiny]{todonotes}

\icmltitlerunning{Latent State Utilization in Probabilistic Circuits}

\begin{document}

\twocolumn[
  \icmltitle{Latent State Utilization in Probabilistic Circuits}



  \icmlsetsymbol{equal}{*}

  \begin{icmlauthorlist}
    \icmlauthor{Zilei Shao}{ucla}
    \icmlauthor{Anji Liu}{nus}
  \end{icmlauthorlist}

  \icmlaffiliation{ucla}{University of California, Los Angeles}
  \icmlaffiliation{nus}{School of Computing, National University of Singapore}

  \icmlcorrespondingauthor{Anji Liu}{anjiliu@nus.edu.sg}

  \icmlkeywords{Probabilistic Circuits, Parameter Learning, Expectation-Maximization}

  \vskip 0.3in
]



\printAffiliationsAndNotice{}  

\begin{abstract}
Probabilistic circuits (PCs) are generative models that support efficient probabilistic inference.
Adding latent states to a PC increases its representational capacity, but yields diminishing gains in likelihood.
We use mutual information to examine how the dependence represented by a trained PC changes with latent size.
Experiments with Hidden Chow-Liu Trees (HCLTs) on ImageNet32 show that learned dependence grows slowly and remains well below both the latent-state upper bound and estimates of dependence in the data.
To understand how learning uses these states, we examine how strongly each sample favors particular latent states and how broadly states are used across the dataset.
Tracking these measurements during HCLT training distinguishes rapid early posterior concentration from the more gradual stabilization of sample-dependent state preferences.
Early measurements also guide initialization selection in HCLTs and hidden Markov models (HMMs), as well as tree selection in HCLTs.
The resulting choices improve the ultimate performance across the tested latent sizes, providing a practical benefit from our study of latent-state utilization.
\end{abstract}

\section{Introduction}
Tractable probabilistic models (TPMs) support efficient probabilistic inference, making them useful for constrained decoding \citep{zhang2023tractable, zhang2024adaptable, weng2025trace, dang2026mitigatingbiaslocallyconstrained}, speculative decoding \citep{grivas2026fastexpressivemultibyteprediction}, and modeling dependencies in diffusion language models \citep{li2026breaking}.
Probabilistic circuits (PCs) provide a general framework for tractable probabilistic modeling by recursively composing distributions through sum and product operations.
Recent advances in structure learning and parameter learning have improved their ability to fit data \citep{wagner2026online,liu2026rethinking}.
However, achieving sustained improvements from increasing model size remains a challenge.
Previous work reported that likelihood gains plateau under conventional parameter-learning methods \citep{liu2023scaling}, and our experiments show diminishing returns even with Anemone, a recent mini-batch expectation-maximization (EM) algorithm.
These findings motivate exploring how effectively learning uses the representational capacity from scaling.

The interpretation of PCs as latent-variable models \citep{peharz2016latent} provides a way to examine how learning uses this additional capacity.
Adding latent states increases a model's capacity, but how those states are used also matters.
Prior work on semantically informed initialization has shown that guiding this organization can improve subsequent learning \citep{liu2023scaling}.
We study this process of \emph{latent-state utilization} from an information-theoretic perspective. We use mutual information to examine how much dependence between variables a trained PC represents.
Our experiments with Hidden Chow-Liu Trees (HCLTs) on ImageNet32 show that learned dependence grows slowly with the number of latent states and remains well below both the latent-state upper bound and estimates of dependence in the data, even in the largest models tested.
These findings motivate examining how learning organizes the available states.

Our information-theoretic analysis relates learned dependence to per-sample posterior concentration, which we call \emph{selectivity}, and aggregate state usage across samples.
We measure these properties through flow entropy and
usage entropy, respectively.
We track these measurements during training to examine how latent-state differentiation develops.
The results distinguish rapid early posterior concentration from the more gradual stabilization of sample-dependent state preferences.
This distinction motivates asking whether early measurements can already indicate which models will achieve better likelihood after training.

These measurements guide early initialization selection in HCLTs and hidden Markov models (HMMs), as well as tree selection in HCLTs.
The resulting choices improve PC performance across the tested latent sizes, providing a practical payoff from our information-theoretic study of latent-state differentiation.
\vspace{-0.3em}
\section{Preliminaries}

A probabilistic circuit (PC) represents a  distribution using a rooted directed acyclic graph of input, sum, and product nodes~\citep{ProbCirc20,poon2011sum}. Input nodes define univariate distributions over variables $X \in \X$. Sum and product nodes recursively combine these distributions into a joint distribution over $\X$. Formally, for an assignment $\x$ to $\X$, each node computes
\begin{equation}
    p_n(\x)=
    \begin{cases}
        f_n(\x),
            & n\text{ is an input node},\\
        \prod_{c\in\ch(n)}p_c(\x),
            & n\text{ is a product node},\\
        \sum_{c\in\ch(n)}\params_{n,c} \cdot p_c(\x),
            & n\text{ is a sum node},
    \end{cases}
    \label{eq:bg-pc}
    \notag
\end{equation}
\noindent where $\ch(n)$ denotes the children of $n$ and $f_n$ is a univariate distribution (e.g., categorical or Bernoulli). The sum node parameters $\{\params_{n,c}\}_{n,c}$ satisfy $\params_{n,c}\geq0$ and $\sum_{c\in\ch(n)}\params_{n,c}=1$ ($\forall \text{~sum~node~} n$). 
For positive weights, we also use the log-parameters $\phi_{n,c}=\log\params_{n,c}$ and denote their collection by $\phi$.

Appropriate structural constraints allow PCs to support tractable probabilistic inference. In particular, smoothness and decomposability  are sufficient for exact and efficient computation of marginal and conditional probabilities \citep{pmlr-v38-peharz15}. These constraints are expressed through the scope $\operatorname{Var}(n)$ of a node $n$, defined as the set of variables associated with input nodes in the subcircuit rooted at $n$.
\begin{definition}[Smoothness and decomposability]
\label{def:smooth-decomposable}
A PC is \emph{smooth} if, for every sum node $n$, all children $c_1,c_2\in\ch(n)$ have the same scope,
$\operatorname{Var}(c_1)=\operatorname{Var}(c_2)$.
A PC is \emph{decomposable} if, for every product node $n$, all distinct children $c_1,c_2\in\ch(n)$ have disjoint scopes,
$\operatorname{Var}(c_1)\cap\operatorname{Var}(c_2)=\emptyset.
$
\end{definition}
These conditions result in a circuit that is multilinear in its input-node values, allowing efficient marginalization \citep{3702676.3702696}. We assume both properties throughout.

\section{Diminishing Returns from Scaling}
\label{sec:scaling}

Recent advances have improved PC learning through both structure learning and parameter learning.
Structure-learning methods adapt circuit structures to data, including through online structural updates \citep{wagner2026online}.
Parameter-learning methods focus on optimizing a given circuit. For example, Anemone regularizes distribution changes during mini-batch EM to limit overfitting to individual batches \citep{liu2026rethinking}.

Despite these advances, increasing model size can still produce diminishing improvements in model fit. Previous work reported that the performance of PCs trained with conventional parameter-learning methods can plateau as their size increases~\citep{liu2023scaling}. We observe diminishing returns with Anemone as well.
For Hidden Chow-Liu Trees \footnote{See Appendix \ref{app:hclt} for details.} with a fixed tree topology, successive doublings of the number of latent states produce smaller gains in validation log-likelihood under a common training budget (\cref{fig:mi-scaling}).
To understand these diminishing returns, we examine how scaling changes a PC's ability to model dependencies between variables.

\begin{figure}[t]
    \centering
    \includegraphics[width=\linewidth]{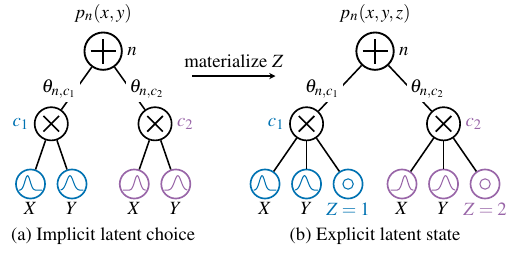}
    \caption{\textbf{Materializing a latent variable.}
    (a) A sum node mixes two components, each factorized over $X$ and $Y$.
    (b) Indicator inputs $\mathbb{I}[Z=1]$ and $\mathbb{I}[Z=2]$
    make the selected component explicit.
    Marginalizing $Z$ recovers the original circuit.}
    \label{fig:latent-materialization}
\end{figure}

\subsection{Latent size and learned dependence}

\begin{figure}[t]
    \centering
    \includegraphics[width=\linewidth]{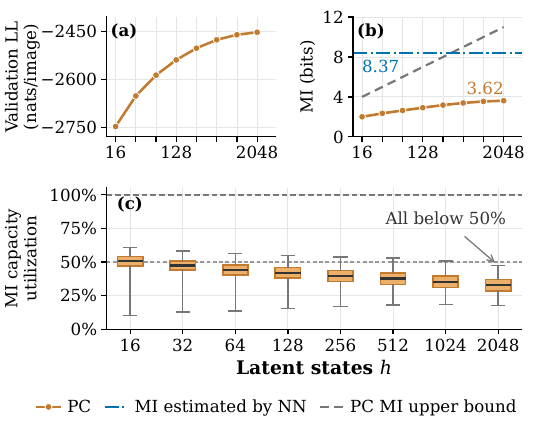}
    \caption{\textbf{Increasing latent size yields diminishing likelihood gains and lower MI capacity utilization.}
    HCLTs on ImageNet32 share a tree topology, initialization scheme, and 180-epoch Anemone training recipe.
    \textbf{(a)} PC validation log-likelihood in nats per image.
    \textbf{(b)} Mean learned MI (orange), the PC MI upper bound $\log_2 h$ (gray dashed), and the mean neural InfoNCE lower-bound estimate (blue dash-dot), in bits.
    The neural estimate uses $N_c=4096$ and is independent of $h$.
    \textbf{(c)} MI capacity utilization,
    $I_{p_{\params}}(\X;\Y)/\log_2 h$.
    All MI measurements use the same set of non-root latent-induced partitions.
    Boxes show medians and quartiles, with full-range whiskers.}
    \label{fig:mi-scaling}
\end{figure}

PCs can be interpreted as hierarchical latent-variable
models~\citep{peharz2016latent}. Each sum node represents a mixture of its child distributions.
For a sum node with $h$ children, the choice of component can be represented by a discrete latent variable $Z$, where $Z=z$ corresponds to selecting child $c_z$. As illustrated in \cref{fig:latent-materialization},
this choice can be made explicit by associating each child with an indicator for its latent state. Marginalizing out $Z$ recovers the original mixture. 

This latent choice allows a mixture to represent dependencies that are absent within its components.
When the child distributions factorize over the same two disjoint variable blocks $\X$ and $\Y$, the blocks are independent within each component.
Their shared component choice can nevertheless induce dependence between them.
Increasing the number of latent states provides additional mixture components for modeling this dependence.

Formally, let $p_n$ denote the normalized distribution represented by sum node $n$.
Suppose that $\X$ and $\Y$ partition its scope and that each child distribution factorizes over these blocks.
Let $\x$ and $\y$ denote assignments to the two blocks. The local joint distribution can be written as
\begin{equation}
    p_n(\x,\y)
    =
    \sum_{z=1}^{h}
    p_n(z)\,p_n(\x\mid z)\,p_n(\y\mid z),
    \label{eq:latent-mixture}
\end{equation}
where $p_n(z)=p_n(Z=z)=\params_{n,c_z}$. The mixture weights and conditional distributions depend on $\params$. We omit this from their notation for readability.

\textbf{Dependence contributes to the whole-model likelihood.}
The contribution of dependence can be expressed in log-likelihood terms.
For the local mixture, consider replacing $p_n(\x,\y)$ with the independent distribution $p_n(\x)p_n(\y)$, which preserves both marginals but removes their dependence. On samples drawn from $p_n$, the expected log-likelihood advantage of the joint distribution is
\begin{equation}
\begin{aligned}
    &\mathbb{E}_{(\x,\y)\sim p_n}
    \left[
        \log_2
        \frac{p_n(\x,\y)}{p_n(\x)p_n(\y)}
    \right]\\
    &\qquad=
    \mathbb{E}_{(\x,\y)\sim p_n}
    \left[
        \log_2\frac{p_n(\y\mid\x)}{p_n(\y)}
    \right]
    = I_{p_n}(\X;\Y).
\end{aligned}
\label{eq:mi-likelihood-gain}
\end{equation}

This relationship extends to a decomposition of the whole-model likelihood.
Let $p_{\params}$ denote the full PC distribution over the individual observed variables $X_1,\ldots,X_d$.
Starting with all observed variables, recursively split each block into two disjoint blocks until each contains one variable.
These splits form a nested hierarchy, with each subsequent split subdividing a block from an earlier split.
Applying the identity in \cref{eq:mi-likelihood-gain} recursively to marginals of $p_{\params}$ gives
\begin{equation}
    \mathbb E_{p_{\params}}
    \left[
        \log_2
        \frac{p_{\params}(X_1,\ldots,X_d)}
             {\prod_{j=1}^{d}p_{\params}(X_j)}
    \right]
    =
    \sum_{(\mathbf A,\mathbf B)}
    I_{p_{\params}}(\mathbf A;\mathbf B).
    \label{eq:hierarchical-likelihood}
\end{equation}
Here, the sum runs over the splits $(\mathbf A,\mathbf B)$ in this hierarchy. All expectations and mutual information terms are evaluated under the full PC distribution, not the data distribution.
The block distributions are marginals of $p_{\params}$ and
may differ from the distributions $p_n$ represented by
internal nodes.
The identity follows from cancellation of the intermediate block entropies, also known as the grouping property of total correlation \citep{kraskov2008micmutualinformationbased}.

The expected log-likelihood advantage over the fully factorized distribution therefore decomposes into dependence contributions across a hierarchy of variable partitions.
Any partition of all observed variables can serve as the first split, and the remaining terms account for dependence within its two blocks.
This gives a whole-model interpretation to our analysis of individual partitions.

\textbf{Latent size limits representable dependence.}
Since $\X$ and $\Y$ are conditionally independent given $Z$, the data-processing inequality gives
\begin{equation}
\begin{aligned}
    I_{p_n}(\X;\Y)
    &\le I_{p_n}(\X;Z)
    \le H_{p_n}(Z)\\
    &= -\sum_{z=1}^{h}
       \params_{n,c_z}\log_2\params_{n,c_z}
    \le \log_2 h.
\end{aligned}
\label{eq:latent-mi-bound}
\end{equation}
The bound is reached when the mixture weights are uniform and the selected latent state can be determined from each block separately. Doubling the number of latent states from $h$ to $2h$ increases this mutual information upper bound by one bit.
The same argument applies under the full-model distribution $p_{\params}$ whenever an $h$-state latent variable $Z$ makes the two blocks conditionally independent.
In that case, the latent probabilities are $p_{\params}(Z=z)$ rather than the local weights $\params_{n,c_z}$.

\textbf{Learned dependence grows slowly with latent size.}
However, this increase in representational capacity is not matched by a comparable increase in learned dependence.
Our HCLT experiments on $16\times16$ crops from ImageNet32 use a fixed tree topology, initialization scheme, and Anemone training recipe across latent sizes.
Each partition separates the observed descendants of a non-root latent variable from the remaining variables. The two blocks are conditionally independent given that latent variable, as illustrated in \cref{fig:app-hclt-latent-tree} in the appendix.

Across all partitions, we observe an average MI gain of only $0.23$ bits per doubling from $h=16$ to $h=2048$ (\cref{fig:mi-scaling}b).
At $h=2048$, every measured partition achieves less than half the upper bound (\cref{fig:mi-scaling}c).

Such limited growth in learned dependence is not explained by the model having already matched the dependence present in the data.
For the same partitions, we estimate lower bounds on $I_P(\X;\Y)$ using neural critics evaluated with InfoNCE, where $P$ denotes the data distribution.
The mean lower-bound estimate is $8.37$ bits, compared with $3.62$ bits of mean learned MI at $h=2048$. Appendix~\ref{app:mi-scaling} provides the experimental setup and MI estimation procedure. This gap is more pronounced for larger scopes (Appendix~\ref{app:mi-scope-size}).

Given that the learned mutual information falls below both the data-MI lower bound and the model's capacity, we investigate whether this gap reflects a limitation of parameter learning, focusing on how latent states are organized to capture dependencies in the data.

\begin{figure*}[t]
    \centering
    \captionsetup[subfigure]{labelfont=bf}

    \begin{subfigure}[t]{\linewidth}
        \centering
        \includegraphics[width=0.95\linewidth]{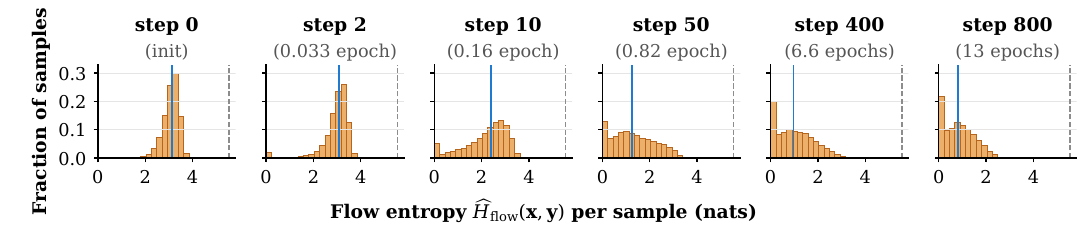}
        \vspace{-0.4em}
        \caption{\textbf{Per-sample flow entropy decreases rapidly during early updates, followed by slower concentration.}
        Histograms show the distribution of posterior entropy across samples for one latent variable.}
        \label{fig:dynamics-entropy}
    \end{subfigure}

    \vspace{0.6em}
    \begin{subfigure}[t]{\linewidth}
        \centering
        \includegraphics[width=0.95\linewidth]{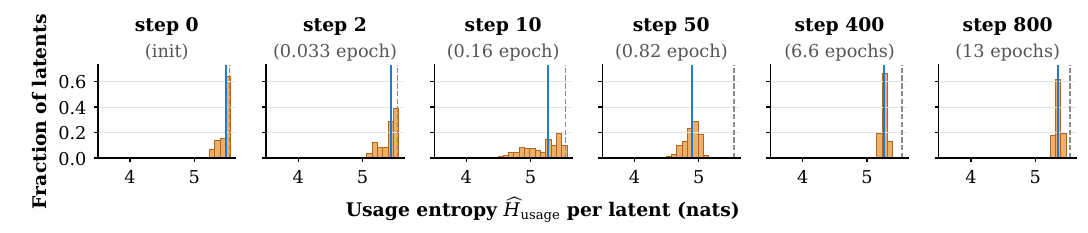}
        \vspace{-0.4em}
        \caption{\textbf{Aggregate state usage becomes less balanced early in training and more balanced later.}
        Histograms show the distribution of usage entropy across sum-node groups.}
        \label{fig:dynamics-usage}
    \end{subfigure}

    \caption{Posterior concentration and aggregate state usage during
    Anemone training (\emph{Unif-pert(16)}, $h=256$), evaluated on a fixed
    set of samples. Blue lines mark medians and dashed lines indicate
    the maximum entropy attained by uniform state probabilities.}
    \label{fig:dynamics}
    \vspace{-0.4em}
\end{figure*}

\subsection{Initialization and latent-state differentiation}
Capturing dependence through \cref{eq:latent-mixture} requires the component distributions to differ between latent states. Previous work illustrates how initialization can encourage this differentiation. Latent variable distillation (LVD) uses deep generative teachers to provide semantically informed latent assignments, improving model fit after subsequent parameter learning \citep{liu2023scaling}.
This motivates studying how initialization shapes the sample-dependent assignments that develop during training.

We begin with the opposite limiting case, in which every latent state starts with the same mixture weight and identically initialized component parameters.
The posterior is then uniform for every sample, so the observed sample provides no information about which state generated it.
The following theorem shows that standard parameter updates cannot break this symmetry on their own.

\begin{theorem}[Preservation of latent-state symmetry,
informal]
\label{thm:latent-symmetry}
For the mixture in \cref{eq:latent-mixture}, suppose all latent states have uniform mixture weights and identical component parameterizations and initial parameters.
In exact arithmetic, EM-based and gradient-based methods preserve this symmetry when they apply the same update rule to every latent state using the same training data, with identical update settings and symmetrically initialized
optimizer states.
Consequently, for every iteration $t$,
\begin{equation}
\begin{aligned}
    p_n^{(t)}(z\mid\x,\y)
    =p_n^{(t)}(z)=\frac1h, \quad
    I_{p_n^{(t)}}(\X;\Y)=0.
\end{aligned}
\label{eq:symmetry-preservation}
\end{equation}
\end{theorem}

The full statement and proof are provided in
Appendix~\ref{app:latent-symmetry}.

In this case, every component receives the same posterior-weighted training data under EM-based updates and the same corresponding gradients under gradient-based updates.
The components therefore change together.
They can improve their marginal fit, but their mixture remains a product distribution over $\X$ and $\Y$.
The dependence contribution in \cref{eq:mi-likelihood-gain} consequently remains zero.
Using random mini-batches does not change this conclusion, since all components are updated on the same samples.

Learning cross-block dependence therefore requires the latent states to depart from this symmetric configuration and develop sample-dependent roles. LVD encourages such differentiation through teacher-derived latent assignments.
Its success motivates examining how initialization shapes the use of latent states during subsequent parameter learning.

\section{Characterizing Latent-State Utilization}
\label{sec:latent-utilization}

\subsection{Aggregate Usage and Posterior Ambiguity}
\label{sec:aggregate_usage_posterior_ambiguity}

To characterize the effects of initialization, we examine the latent-state posteriors $p(z\mid\x,\y)$.
Departing from uniform posteriors does not by itself imply useful differentiation.
Assignments may concentrate on the same few states across samples, or distinguish patterns specific to only one variable block.

We therefore return to mutual information to examine how broadly latent states are used across samples, how strongly each sample favors particular states, and how these
assignments contribute to dependence between $X$ and $Y$.
The following formalizes this relationship.

\begin{proposition}[Latent-state information decomposition]
\label{prop:latent-information-budget}
Let $Z$ have $h$ states and satisfy $\X\perp \Y\mid Z$ under the model distribution. Then
\begin{equation}
\begin{aligned}
    I(\X;\Y)
    ={}&
    \underbrace{H(Z)}_{
        \substack{\text{aggregate}\\\text{usage}}
    }
    -
    \underbrace{H(Z\mid \X,\Y)}_{
        \substack{\text{posterior}\\\text{ambiguity}}
    }\\
    &-
    \underbrace{
        \bigl[I(\X;Z\mid \Y)+I(\Y;Z\mid \X)\bigr]
    }_{\text{block-specific information}}.
\end{aligned}
\label{eq:latent-information-budget}
\end{equation}
All quantities are evaluated under the model distribution and measured in bits.
\end{proposition}

The derivation is provided in Appendix~\ref{app:latent-information-budget}. Aggregate usage $H(Z)$ measures how broadly states participate across samples.
Posterior ambiguity $H(Z\mid\X,\Y)$ measures the uncertainty
about the latent state after observing a sample.
Lower posterior ambiguity corresponds to greater selectivity.
Their difference is $I(Z;\X,\Y)$, the information shared between the latent state and the sample.
The block-specific terms account for the part of this information that does not contribute to cross-block dependence. For example, $\X$ may distinguish latent states that $\Y$ cannot distinguish.

We first focus on the two entropy terms to characterize how latent states differentiate across samples.

The symmetric initialization in \cref{thm:latent-symmetry} illustrates why aggregate usage alone is insufficient.
Every sample has the same uniform posterior, giving $H(Z)=H(Z\mid \X,\Y)=\log_2 h$. All states participate equally, but the posterior does not distinguish samples.
At the opposite extreme, every sample assigns all posterior mass to the same state. Both entropies are then zero. The posteriors are fully concentrated, but the assignments again do not distinguish samples.
In both cases, $I(Z;\X,\Y)=0$.

Sample-dependent differentiation is instead characterized by broad aggregate usage together with low posterior ambiguity.
Different samples then favor different states, rather than sharing a uniform posterior or concentrating on a single state.
We examine per-sample selectivity jointly with aggregate usage throughout training.
We next relate both measurements to the node flows computed
during EM-based learning.

\begin{figure*}[t]
    \centering
    \captionsetup[subfigure]{labelfont=bf}

    \begin{subfigure}[t]{\linewidth}
        \centering
        \includegraphics[width=0.9\linewidth]{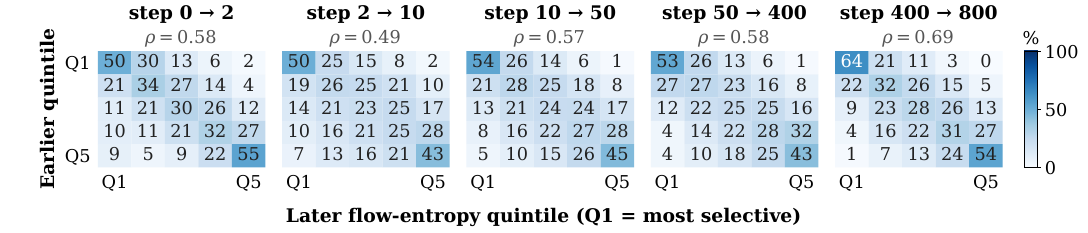}
        \vspace{-0.4em}
        \caption{\textbf{Relative selectivity becomes more stable later in training.}
        Rows and columns represent earlier and later flow-entropy quintiles.
        Each row shows how images move between quintiles.
        Greater diagonal concentration indicates more stable rankings.
        $\rho$ is the Spearman correlation of per-sample entropies.}
        \label{fig:dynamics-transition}
    \end{subfigure}

    \vspace{0.6em}
    \begin{subfigure}[t]{\linewidth}
        \centering
        \includegraphics[width=0.9\linewidth]{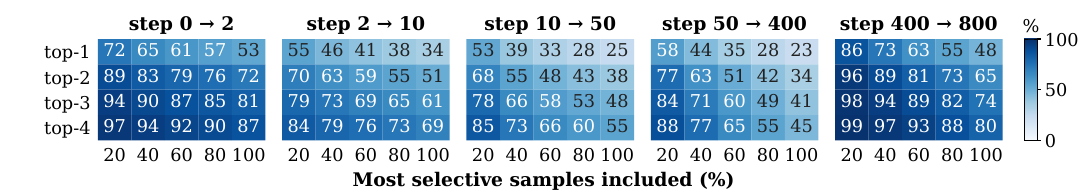}
        \vspace{-0.4em}
        \caption{\textbf{Preferred states become more persistent, especially for selective samples.}
        Columns include progressively larger subsets, from the lowest-entropy
        fifth at the earlier checkpoint to all images.
        Entries show how often the earlier highest-posterior state
        remains among the later top-$k$ states.}
        \label{fig:dynamics-persistence}
    \end{subfigure}

    \caption{Stabilization of sample-dependent state preferences during
    Anemone training (\emph{Unif-pert(16)}, $h=256$).
    Both panels follow the same latent variable and sample set.
    Cell values are percentages.}
    \label{fig:dynamics-stability}
    \vspace{-0.4em}
\end{figure*}

\subsection{Measuring State Usage and Selectivity}

To measure aggregate usage and posterior ambiguity, we need $p_{\params}(z\mid\x,\y)$ for each sample.
In a smooth, decomposable PC, a generative traversal starts at the root, selects one child at each visited sum node, and follows all children of each visited product node. Given a sample $(\x,\y)$, a node's flow is the probability that this traversal visits the node. Flows can be computed by backpropagation \citep{liu2026rethinking}.

Consider a state-indexed group $\mathcal G_Z=\{n_1,\ldots,n_h\}$ such that every traversal visits exactly one node in the group, with visiting $n_z$ corresponding to selecting $Z=z$.
The node flows give the latent-state posterior,
\begin{equation}
    F_\phi^{(\x,\y)}(n_z)
    = p_{\params}(z\mid\x,\y),
    \quad
    \sum_{z=1}^{h}F_\phi^{(\x,\y)}(n_z)=1,
    \label{eq:flow-posterior}
\end{equation}
where $\phi$ denotes the circuit's log-parameters.

We evaluate these flows on a fixed probe set $\mathcal D$. For a fixed group, let $\widehat H_{\mathrm{flow}}(\x,\y)$ denote a sample's posterior entropy.
The \emph{flow entropy} averages these per-sample entropies over $\mathcal D$,
\begin{equation}
\begin{aligned}
    \widehat H_{\mathrm{flow}}^{\mathcal D}(\mathcal G_Z)
    :=
    -\frac{1}{|\mathcal D|} 
    \sum_{(\x,\y)\in\mathcal D}\sum_{z=1}^{h}
    F_\phi^{(\x,\y)}(n_z)
    \ln F_\phi^{(\x,\y)}(n_z).
    \notag
\end{aligned}
\label{eq:empirical-flow-entropy}
\end{equation}
Lower flow entropy means that individual samples favor fewer states, that is, greater per-sample selectivity.

The \emph{usage entropy} instead averages the posteriors before taking entropy.
Writing
$F_\phi^{\mathcal D}(n_z)
=|\mathcal D|^{-1}\sum_{(\x,\y)\in\mathcal D}
F_\phi^{(\x,\y)}(n_z)$,
we define
\begin{equation}
    \widehat H_{\mathrm{usage}}^{\mathcal D}(\mathcal G_Z)
    :=
    -\sum_{z=1}^{h}
    F_\phi^{\mathcal D}(n_z)
    \ln F_\phi^{\mathcal D}(n_z).
    \label{eq:empirical-usage-entropy}
    \notag
\end{equation}
Higher usage entropy means that aggregate posterior mass is more evenly distributed across states.

\begin{table*}[t]
    \centering
    \caption{\textbf{Tree and initialization selection improve likelihood.}
    Final validation LL (nats/image) for Anemone-trained HCLTs.
    Details about initialization are in Appendix \ref{app:initialization}. 
    Tree gains compare matching initializations across trees.
    LVD and initialization gains are relative to equal weights on $t_1$, and combined gains to equal weights on $t_0$. Gains are calculated before rounding.
    Bold and underline mark the best and second-best.\\}
    \vspace{-0.8em}
    \label{tab:initialization-scaling}
    \begingroup
    \small
    \setlength{\tabcolsep}{3pt}
    \renewcommand{\arraystretch}{1.0}
    \resizebox{0.86\textwidth}{!}{%
    \begin{tabular}{l *{7}{c}}
        \toprule
        & \multicolumn{7}{c}{\textbf{Latent states $h$}} \\
        \cmidrule(lr){2-8}
        \textbf{Initialization}
        & \textbf{16} & \textbf{32} & \textbf{64} & \textbf{128}
        & \textbf{256} & \textbf{512} & \textbf{1024} \\
        \midrule

        \multicolumn{8}{l}{\textbf{Baseline tree ($t_0$)}} \\

        Equal weights
        & $-2753.29$ & $-2656.26$ & $-2589.63$ & $-2542.23$
        & $-2506.22$ & $-2480.20$ & $-2467.50$ \\
        \addlinespace[1pt]

        \rowcolor{baselinegray}
        Unif-pert($64$)
        & $-2750.22$ & $-2654.17$ & $-2588.84$ & $-2540.60$
        & $-2504.24$ & $-2477.87$ & $-2462.21$ \\
        \midrule

        \multicolumn{8}{l}{\textbf{Selected tree ($t_1$)}} \\

        Equal weights
        & $-2749.93$ & $-2652.63$
        & $-2585.93$ & $-2538.00$
        & $-2501.83$ & $-2475.86$
        & $-2464.01$ \\
        \quad\footnotesize\textit{Tree selection gain}
        & \footnotesize\textcolor{posgreen}{$+3.36$} & \footnotesize\textcolor{posgreen}{$+3.63$}
        & \footnotesize\textcolor{posgreen}{$+3.70$} & \footnotesize\textcolor{posgreen}{$+4.23$}
        & \footnotesize\textcolor{posgreen}{$+4.39$} & \footnotesize\textcolor{posgreen}{$+4.35$}
        & \footnotesize\textcolor{posgreen}{$+3.49$} \\
        \addlinespace[1pt]

        \rowcolor{baselinegray}
        Unif-pert($64$)
        & \underline{$-2747.23$} & \underline{$-2651.58$}
        & \underline{$-2585.54$} & $-2536.70$
        & $-2499.78$ & \underline{$-2473.56$}
        & \underline{$-2458.67$} \\
        \quad\footnotesize\textit{Tree selection gain}
        & \footnotesize\textcolor{posgreen}{$+2.99$} & \footnotesize\textcolor{posgreen}{$+2.59$}
        & \footnotesize\textcolor{posgreen}{$+3.30$} & \footnotesize\textcolor{posgreen}{$+3.89$}
        & \footnotesize\textcolor{posgreen}{$+4.46$} & \footnotesize\textcolor{posgreen}{$+4.31$}
        & \footnotesize\textcolor{posgreen}{$+3.54$} \\
        \addlinespace[1pt]

        \rowcolor{baselinegray}
        LVD
        & $-2750.88$ & $-2653.94$
        & $-2586.06$ & \underline{$-2536.40$}
        & \underline{$-2499.31$} & $-2474.89$
        & $-2461.37$ \\
        \quad\footnotesize\textit{LVD gain}
        & $-0.95$ & $-1.31$
        & $-0.12$ & \footnotesize\textcolor{posgreen}{$+1.60$}
        & \footnotesize\textcolor{posgreen}{$+2.52$} & \footnotesize\textcolor{posgreen}{$+0.97$}
        & \footnotesize\textcolor{posgreen}{$+2.64$} \\
        \addlinespace[1pt]

        \rowcolor{lightgreen}
        Selected initialization
        & $\mathbf{-2743.60}$ & $\mathbf{-2646.18}$
        & $\mathbf{-2580.41}$ & $\mathbf{-2533.30}$
        & $\mathbf{-2497.41}$ & $\mathbf{-2471.54}$
        & $\mathbf{-2455.89}$ \\
        \quad\footnotesize\textit{Initialization gain}
        & \footnotesize\textcolor{posgreen}{$+6.33$} & \footnotesize\textcolor{posgreen}{$+6.45$}
        & \footnotesize\textcolor{posgreen}{$+5.53$} & \footnotesize\textcolor{posgreen}{$+4.70$}
        & \footnotesize\textcolor{posgreen}{$+4.42$} & \footnotesize\textcolor{posgreen}{$+4.31$}
        & \footnotesize\textcolor{posgreen}{$+8.13$} \\
        \quad\footnotesize\textit{Combined gain}
        & \footnotesize\textcolor{posgreen}{$+9.69$} & \footnotesize\textcolor{posgreen}{$+10.08$}
        & \footnotesize\textcolor{posgreen}{$+9.22$} & \footnotesize\textcolor{posgreen}{$+8.93$}
        & \footnotesize\textcolor{posgreen}{$+8.81$} & \footnotesize\textcolor{posgreen}{$+8.66$}
        & \footnotesize\textcolor{posgreen}{$+11.62$} \\
        \bottomrule
    \end{tabular}%
    }
    \endgroup
    \vspace{-1.0em}
\end{table*}

Both measurements are reported in nats. On model-generated samples, they estimate $H(Z\mid\X,\Y)$ and $H(Z)$, respectively. Dividing by $\ln 2$ converts them to bits.
Our experiments instead use a fixed real-data probe set, so the measurements describe state usage and selectivity on the observed data.

On this probe set, the difference $\widehat H_{\mathrm{usage}}-\widehat H_{\mathrm{flow}}$ equals the average KL divergence between each sample's posterior and the aggregate posterior.
It therefore measures how much the posteriors vary across samples. Recall the two limiting cases in \cref{sec:aggregate_usage_posterior_ambiguity}. Uniform posteriors give broad aggregate usage without per-sample selectivity, whereas concentrating every sample's posterior on the same state gives selectivity without broad usage.
In both cases, the posteriors do not distinguish samples, so usage entropy equals flow entropy and their difference is zero.

Jointly tracking the two entropies allows us to examine whether posterior concentration develops alongside sample-dependent state preferences.

\vspace{-0.2em}
\section{Learning Dynamics of Selectivity}

We track flow entropy, usage entropy, and each sample's highest-posterior state during training of an HCLT on ImageNet32, with a fixed tree and $h=256$ latent states. We use uniform-noise initialization (\emph{Unif-pert(16)}).
Starting from equal weights, we add independent noise drawn uniformly from $[-16,0]$ to the sum and input log-weights, then normalize each log-weight vector with softmax (Appendix \ref{app:initialization}).
On a fixed probe set, we track per-sample posteriors for one fixed sum-node group and usage entropy across all non-root latent groups.

\textbf{Posteriors sharpen rapidly.}
Median per-sample flow entropy decreases from $3.14$ to $1.23$ nats within the first $50$ updates, corresponding to less than one epoch, and reaches $0.82$ nats by step $800$ (\cref{fig:dynamics-entropy}).
Most posterior sharpening therefore occurs early, followed by slower concentration.

\textbf{Aggregate usage recovers after an initial decline.}
Usage entropy initially decreases across sum-node groups, indicating that aggregate posterior mass becomes less evenly distributed across states (\cref{fig:dynamics-usage}).
It later increases even as per-sample flow entropy continues to decrease.
Later training thus combines concentrated individual posteriors with broader aggregate state usage.

\textbf{State preferences stabilize more gradually.}
Despite early posterior sharpening, only $23\%$ of samples keep their highest-posterior state between steps $50$ and $400$, increasing to $48\%$ between steps $400$ and $800$ (\cref{fig:dynamics-persistence}).
The rank correlation of per-sample entropies also increases from $0.58$ to $0.69$ across these intervals (\cref{fig:dynamics-transition}).
Persistence is substantially higher for selective samples.
Among the most selective fifth at step $400$, $86\%$ keep their highest-posterior state at step $800$.
Posterior concentration therefore precedes the more gradual stabilization of state preferences, with lower-entropy samples showing greater persistence.

Appendix~\ref{app:initialization-dynamics} extends the dynamics analysis to LVD and Dirichlet initialization.

Together, these findings distinguish rapid posterior
sharpening from the later recovery in aggregate usage
and increasing persistence of state preferences.
Flow entropy and usage entropy jointly distinguish
selective assignments spread across states from
concentration on the same few states.

\section{Early Latent Utilization for Model Selection}
\label{sec:model-selection}

The preceding section characterizes how selectivity, aggregate usage, and the persistence of state preferences develop within a training run. We now ask whether early differences in selectivity and aggregate usage across candidate models help identify initializations and tree structures that achieve better likelihood after training.
The rapid posterior sharpening motivates probing initializations after two epochs.

Since low flow entropy can arise when samples concentrate on the same few states, we use the entropy difference as a gate before selecting by flow entropy.
For each candidate, we average both measurements across latent groups and compute $\widehat H_{\mathrm{usage}}-\widehat H_{\mathrm{flow}}$.
Within each pool, we exclude candidates whose entropy difference is below $75\%$ of the pool median and select the remaining candidate with minimum flow entropy.
\vspace{-0.15em}
\subsection{Selecting Initializations}
\begin{figure}[t]
    \centering
    \includegraphics[width=0.95\linewidth]{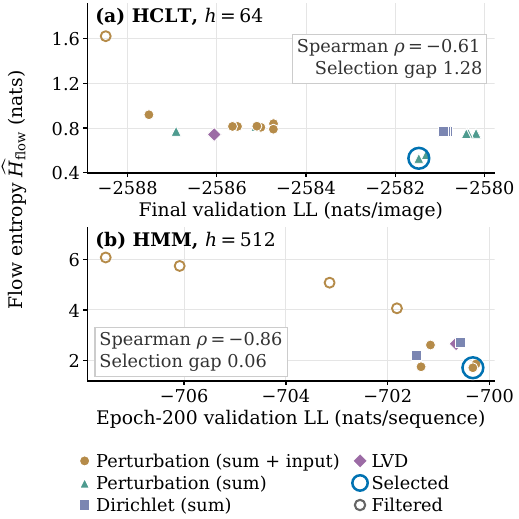}
    \vspace{-0.2em}
    \caption{\textbf{Early flow entropy guides initialization selection.}
    Flow entropy after two epochs is compared with final validation likelihood for HCLTs and epoch-200 likelihood for HMMs.
    Open markers fail the entropy-difference gate, and blue rings mark selected candidates.
    Boxes report Spearman correlation across each full pool and the selected candidate's likelihood shortfall from the best candidate.}
    \vspace{-0.2em}
    \label{fig:init-selection}
\end{figure}

\vspace{-0.15em}
We compare 19 initializations for HCLTs with $h=64$ on a fixed tree and 13 for HMMs with $h=512$.
We apply the selection rule after two epochs of Anemone training for HCLTs and full-batch EM for HMMs.
We compare perturbations of sum parameters alone or together with input parameters, Dirichlet initialization of sum parameters, and LVD (\cref{fig:init-selection}).

Across the full candidate pools, lower early flow entropy is associated with higher later validation likelihood, with Spearman correlations of $-0.61$ for HCLTs and $-0.86$ for HMMs.
The selected candidates finish within $1.28$ nats per image and $0.06$ nats per sequence of the best candidates in their respective pools.
These results support using early posterior selectivity to compare initializations before completing training.

Table~\ref{tab:initialization-scaling} evaluates initialization selection across latent sizes using separate pools of nine candidates per width on tree $t_1$.
The selected initializations improve final likelihood by $4.31$ to $8.13$ nats per image over equal weights and by $2.02$ to $5.40$ nats per image over Unif-pert($64$).
The fixed Unif-pert($64$) initialization also outperforms LVD at five of seven widths, despite using no teacher-derived latent assignments.
The selected initializations outperform both throughout the table.

Comparisons with direct entropy-difference ranking and early validation likelihood appear in Appendices~\ref{app:entropy-difference-selection} and~\ref{app:early-likelihood}, respectively.
\vspace{-0.15em}
\subsection{Selecting Tree Structures}
\begin{figure}[t]
    \centering
    \includegraphics[width=0.95\linewidth]{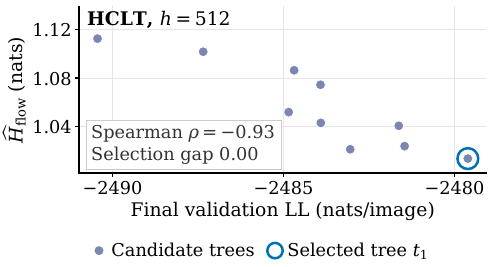}
    \vspace{-0.2em}
    \caption{\textbf{Initial flow entropy selects the best tree in this pool.}
    Ten HCLT trees ($h=512$) share Unif-pert($64$).
    Likelihood is averaged over the last three evaluations of 180-epoch Anemone training.
    The blue ring marks the selected tree.
    The box reports Spearman correlation and the selection gap.}
    \label{fig:tree-selection}
    \vspace{-0.2em}
\end{figure}

\vspace{-0.15em}
We compare ten candidate HCLT trees with $h=512$, using Unif-pert($64$) and the same training recipe.
We apply the same selection rule before any parameter updates.
All ten candidates pass the gate, and initial flow entropy correlates with final validation likelihood at $\rho=-0.93$ (\cref{fig:tree-selection}).
The minimum-entropy tree, $t_1$, achieves the highest final likelihood in the pool.

Using $t_1$ across the widths in Table~\ref{tab:initialization-scaling} improves likelihood over $t_0$ by $3.36$ to $4.39$ nats per image with equal weights and by $2.59$ to $4.46$ with Unif-pert($64$).
Combining tree and initialization selection gives gains of $8.66$ to $11.62$ nats per image over equal weights on $t_0$.
Thus, the table separates improvements from choosing the tree from those achieved by choosing initialization.

Appendix~\ref{app:pic-comparison} additionally compares our patch-based HCLTs with published CP variants \citep{gala2024scaling}.

\section{Related Work}

\paragraph{Circuit structures and parameterizations.}
Monarch matrices reduce the cost of circuit parameterizations \citep{zhang2025monarch}, and tensor-factorization frameworks connect circuit architectures to CP and Tucker decompositions \citep{loconte2025what}. 
Other approaches introduce continuous latent variables through probabilistic integral circuits \citep{gala2024scaling} or adapt deterministic, structured-decomposable circuits through online structural updates \citep{wagner2026online}. 
Einsum Networks combine circuit operations into einsum computations for efficient learning \citep{peharz2020einsum}.
\citet{zhao2026expressivityboundaryprobabilisticcircuits} identify expressivity limitations from probability-space outputs and fixed routing in comparisons between PCs and Transformers.
Our work complements this architectural analysis by examining how learning uses the latent-state capacity of a given circuit. 

\paragraph{Parameter learning and robustness.}
LVD guides latent assignments using a generative teacher \citep{liu2023scaling}, and Anemone regularizes distribution changes during mini-batch EM \citep{liu2026rethinking}.
Sharpness-aware learning penalizes loss curvature \citep{suresh2026tractable}, distinct from posterior concentration, and PeTeR provides data-free post-training robustification \citep{ciotinga2026peterposttrainingrobustificationprobabilistic}.

\paragraph{Task-specific tractable models.}
For controllable language generation, \citet{liu2026the} study alternatives to maximum likelihood, including reverse KL divergence with entropy regularization and sequence log-probability matching.
\citet{yidouweng2026learningtractabledistributionslanguage} instead condition an HMM's initial-state distribution on language-model prefix representations, keeping shared transition and emission parameters and training by conditional maximum likelihood.

\paragraph{Random initialization in circuit implementations.}
PyJuice \citep{pyjuice} and Cirkit \citep{The_APRIL_Lab_cirkit_2024} support random parameter initialization with different sampling distributions and parameterizations.
Our symmetry result explains the need to differentiate otherwise identical components under state-symmetric learning updates.

\paragraph{Tractable operations and inference.}
The compositional atlas characterizes tractable circuit operations for probabilistic and information-theoretic queries \citep{vergari2021compositional}, and geometry-aware PCs study tractability under Voronoi-based gating \citep{sidheekh2026geometryaware}.
Complementary work develops randomized MAP inference with probabilistic approximation guarantees \citep{shorvon2026probably}.

\paragraph{Symmetry and convergence in mixture models.}
\citet{ma2000asymptotic} relate local EM convergence in Gaussian mixtures to component overlap, with faster asymptotic convergence as overlap vanishes.
Our theorem examines exact component symmetry, under which state-symmetric EM-style and gradient-based updates preserve identical components rather than allowing them to differentiate.

\section{Conclusion}
\label{sec:related-conclusion}
Our experiments show that increasing latent size yields diminishing gains in learned dependence, which remains well below the latent-state bound.
Our symmetry analysis complements work linking EM convergence to component overlap in Gaussian mixtures \citep{ma2000asymptotic}, showing that identical components cannot differentiate under state-symmetric updates.
Flow entropy and usage entropy jointly distinguish sample-dependent state preferences from concentration on a few states.
During training, posterior concentration develops before state preferences stabilize.
These measurements also provide an inexpensive probe for initialization and tree selection, improving final likelihood in the candidate pools studied.

\section{Acknowledgments}
This work was funded in part by the National University of Singapore under its Start-up Grant (Award No: SUG-251RES2505); this work was also funded in part by the DARPA CLARA program, and gifts from Cisco Research, Qualcomm, and Amazon.

\bibliographystyle{icml2026}
\bibliography{preprint}

\clearpage
\appendix
\onecolumn

\section{Experiment Details}
\label{app:experimental-details}

\subsection{Hidden Chow--Liu Trees}
\label{app:hclt}

We use hidden Chow--Liu trees (HCLTs), which place a tree structure over latent variables and attach the observed variables as leaves \citep{liu21tractable}.
Starting from a Chow--Liu tree over the observed variables $\X=\{X_1,\ldots,X_d\}$, we replace each $X_j$ with a latent variable $Z_j$, preserving the tree edges.
We then attach $X_j$ as a child of $Z_j$.
Each latent variable has $h$ states.
Thus, dependencies between observed variables are mediated by the latent tree.

Let $r$ index the root and $\operatorname{pa}(j)$ index the parent of a non-root latent variable $Z_j$.
For a joint latent assignment $\mathbf z=(z_1,\ldots,z_d)$, the observed distribution is
\begin{equation}
    p_{\params}(\x)
    =
    \sum_{\mathbf z\in\{1,\ldots,h\}^{d}}
    p(z_r)
    \prod_{j\ne r} p(z_j\mid z_{\operatorname{pa}(j)})
    \prod_{j=1}^{d} p(x_j\mid z_j).
    \label{eq:hclt}
\end{equation}
The factors specify the root distribution, transitions between latent states, and conditional distributions of the observed variables.
We omit their dependence on $\params$ for readability.

\begin{figure*}[t]
    \centering
    \includegraphics[width=0.9\linewidth]{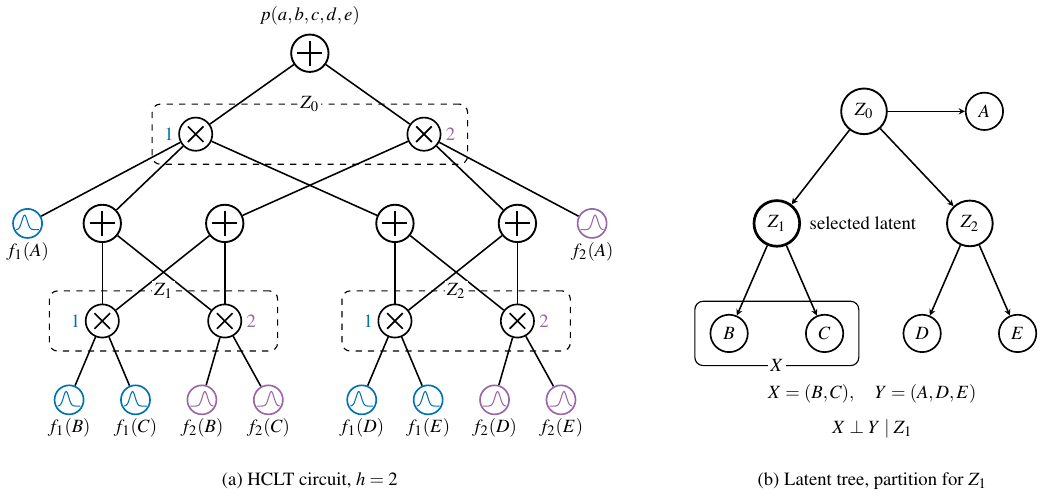}
    \caption{\textbf{An HCLT and its latent-induced variable partition.}
    (a) The circuit sums over latent states, with $h=2$.
    Dashed groups identify the state-specific components of each latent variable.
    (b) In the corresponding latent tree, conditioning on $Z_1$ separates $\X=(B,C)$ from $\Y=(A,D,E)$.
    Terminal latent variables have been absorbed into the input distributions.}
    \label{fig:app-hclt-latent-tree}
\end{figure*}

In our implementation, a terminal latent variable with one observed child is summed out, and the resulting conditional distribution of that observed variable given the parent latent variable is represented by an input node.
This gives the representation illustrated in \cref{fig:app-hclt-latent-tree}.
Here, $Z_0$ has latent children $Z_1$ and $Z_2$ and an observed child $A$.
The observed variables $B,C$ depend on $Z_1$, and $D,E$ depend on $Z_2$.

The circuit in \cref{fig:app-hclt-latent-tree}(a) evaluates this factorization by summing over latent states.
For each state of $Z_0$, a product node combines the conditional distribution of $A$ with the distributions over $(B,C)$ and $(D,E)$.
These two distributions are computed by summing over the states of $Z_1$ and $Z_2$, respectively.
The sum weights encode the latent prior and transition probabilities.
For example, the input labeled $f_z(B)$ represents $p(b\mid Z_1=z)$.

The latent tree also specifies the variable partitions used in our dependence analysis.
For any non-root latent variable $Z$, let $\X$ contain its observed descendants and $\Y$ contain the remaining observed variables.
Conditioning on $Z$ separates the two blocks, giving $\X\perp \Y\mid Z$.
In \cref{fig:app-hclt-latent-tree}(b), choosing $Z_1$ gives $\X=(B,C)$ and $\Y=(A,D,E)$.
Under the full model distribution, each such partition therefore satisfies
\begin{equation}
    p_{\params}(\x,\y)
    =
    \sum_{z=1}^{h}
    p_{\params}(z)\,
    p_{\params}(\x\mid z)\,
    p_{\params}(\y\mid z),
\end{equation}
and consequently $I_{p_{\params}}(\X;\Y)\leq H_{p_{\params}}(Z)\leq\log_2 h$.
The tree determines these partitions, and $h$ controls the number of latent states available to model dependence across each partition.

\subsection{Data and Training}
\label{app:training}

We train HCLTs on aligned $16\times16$ crops of ImageNet32 images.
Each crop is transformed to YCoCg-R and flattened into 768 observed
variables. Categorical input distributions have 512 possible values.
We use no parameter tying and vary the latent state number over $h\in\{16,32,64,128,256,512,1024,2048\}.$
The scaling experiments use the same tree at every width.
Only the number of latent states changes, not the number of latent variables.

\paragraph{Dataset and usage terms.}
We use ImageNet32~\citep{chrabaszcz2017downsampledvariantimagenetalternative}, a $32\times32$ downsampled version of ImageNet~\citep{deng2009imagenet}.
ImageNet provides access for non-commercial research and educational purposes under its terms of access.\footnote{\url{https://www.image-net.org/download}}
ImageNet does not own the copyright of the underlying images.

For the HMM experiments, we use WikiText-103~\citep{merity2017pointer},
a corpus of Wikipedia articles distributed under CC BY-SA 3.0
and GFDL licenses.\footnote{\url{https://huggingface.co/datasets/Salesforce/wikitext}}

\paragraph{Tree construction.}
We construct each tree from the first 20,000 training images in
a fixed file order, sampling one of the four aligned crops from
each image.
Pairwise dependence scores are computed using PyJuice's Gaussian
soft-binning estimator with 64 bins and bandwidth $0.02$.
We construct a maximum-weight spanning tree and root it at a tree center.
The crop choices and resulting topology are fixed across widths.
We verify structural agreement by comparing the complete collection
of latent subtree scopes.

\paragraph{Anemone training.}
We use an effective batch size of $16{,}384$, momentum $0.9$,
and pseudocount $10^{-6}$.
Each epoch processes $10^6$ images and corresponds to approximately
61 updates.
Initialization and training seeds are kept fixed in comparisons
that vary only latent size or tree structure.

The scaling experiments and the $h=512$ tree comparison use
a common 180-epoch budget.
The step size remains at $0.8$ for the first 500 updates, follows
a cosine schedule to $0.4$ at update $10{,}000$, and remains
at $0.4$ thereafter.
Scaling measurements use the checkpoint at epoch 180.

The initialization comparison and cross-width table use a longer
schedule, with 1,000 initial updates at step size $0.8$ and cosine
decay to $0.4$ at update $20{,}000$.
Validation minibatches are divided into two sets,
$\mathcal D_A$ and $\mathcal D_B$, by alternating minibatches.
After the schedule ends, training stops when likelihood on
$\mathcal D_A$ improves by less than $0.1$ nats per image over
50 epochs, or when the run reaches 500 epochs.
We report likelihood on $\mathcal D_B$ at the final evaluation.
Validation crops are fixed across runs and evaluations.

\subsection{Initialization and Posterior Probes}
\label{app:initialization}

\paragraph{Perturbation initialization.}
For each probability vector, we independently sample
log-weights and normalize them with softmax,
\begin{equation}
    q_i^{(0)}
    =
    \frac{\exp(\xi_i)}{\sum_j \exp(\xi_j)},
    \qquad
    \xi_i \overset{\mathrm{i.i.d.}}{\sim}
    \mathcal U[-\lambda,0].
\end{equation}
We denote this uniform-noise initialization by
\emph{Unif-pert($\lambda$)}.

Uniform weights assign equal probability to every entry.
Each sum node therefore assigns probability $1/|\ch(n)|$ to
each child, and each categorical input assigns probability $1/512$
to each value.
For perturbations of sum parameters alone, the input parameters
are kept from the $\lambda=64$ initialization.
Dirichlet initialization independently samples each sum-weight
vector from a symmetric Dirichlet distribution and likewise
leaves the input parameters unchanged.
For large perturbations, parameter blocks containing underflowed
zeros are redrawn with an unnormalized weight floor of $10^{-30}$.

\paragraph{Equal-weight initialization.} The equal-weight baseline uses ordinary floating-point training without enforcing symmetry after initialization. Numerical differences can therefore break the initial symmetry, unlike the exact-arithmetic setting of \cref{thm:latent-symmetry}.

\paragraph{LVD initialization.}
Our HCLT LVD initialization uses a VQ-VAE-2 teacher \citep{vq-vae-2} trained for 20 epochs on the same RGB crops, following prior work \citet{liu2023understandingdistillationprocessdeep}.
We extract features from 200,000 training crops and cluster them into $h$ states using 20 iterations of $k$-means.
The features combine the teacher's decoder parameters at each pixel with embeddings from its patch and neighboring patches.
The feature blocks are standardized and scaled to have equal total variance. Cluster labels are shared by the three channel variables at each pixel.
We estimate latent transition and categorical input distributions from label co-occurrences and conditional value counts, with pseudocount $0.01$.
The tree topology remains fixed.

\paragraph{Posterior probes.}
The learning-dynamics analysis uses $h=256$ and a fixed set of
$24{,}832$ images from $\mathcal D_B$.
We record measurements at updates $0,2,10,50,400,$ and $800$,
with step size $0.8$ throughout this interval.
Per-sample entropy, entropy-quintile transitions, and preferred-state
persistence are measured for the first sum-node group in circuit
order whose scope contains 16 variables.
Usage-entropy histograms summarize the non-root latent groups.

Selection probes use validation samples.
Within each candidate pool, the samples and preprocessing
are fixed across candidates.
Both entropy measurements are averaged equally across
latent groups and used in the selection rule in \cref{sec:model-selection}.

\subsection{Mutual-Information Measurements}
\label{app:mi-scaling}

\paragraph{Learned model dependence.}
We measure MI at all 563 non-root partitions of the fixed scaling tree.
The exact variable sets are shared across widths.
For each trained model, we draw unconditional samples
$(\x_\ell,\y_\ell)\sim p_{\params}$ and compute
\begin{equation}
\begin{aligned}
    \widehat I_{p_{\params}}(X;Y)
    =
    \frac{1}{N\ln 2}
    \sum_{\ell=1}^{N}
    \bigl[
        \ln p_{\params}(\x_\ell,\y_\ell)-\ln p_{\params}(\x_\ell)
        -\ln p_{\params}(\y_\ell)
    \bigr].
\end{aligned}
\label{eq:model-mi-estimator}
\end{equation}
Joint probabilities are evaluated by a circuit forward pass.
Marginal probabilities are obtained by marginalizing the
complementary variable block.
The estimator therefore uses exact circuit probabilities.

We use calibrated PyJuice sampling with the same sampling
procedure and seed at every width.
Each model's complete samples are reused across its partitions.
Each partition is first evaluated on 50,000 samples and extended
to 200,000 if its approximate 95\% Monte Carlo half-width exceeds
$0.02$ bits.
Writing the bracketed log-ratio in
\cref{eq:model-mi-estimator} as $d_\ell$, the standard error is
\begin{equation}
    \widehat{\operatorname{SE}}
    =
    \frac{s_d}{\sqrt N\,\ln 2},
\end{equation}
where $s_d$ is the sample standard deviation of the log-ratios.

Each partition receives equal weight in the mean learned MI.
The capacity fractions are
$\widehat I_{p_{\params}}(X;Y)/\log_2 h$.
Their distributions summarize variation across partitions of
a trained model, rather than variation across training seeds.

\paragraph{Dependence in the data.}
We train separate neural critics for the same 563 partitions
and estimate InfoNCE lower bounds on the data MI.
For a batch of $N_c$ paired samples, the estimate in bits is
\begin{equation}
    \widehat B
    =
    \frac{1}{N_c\ln 2}
    \sum_{i=1}^{N_c}
    \ln
    \frac{N_c\exp s(\x_i,\y_i)}
         {\sum_{j=1}^{N_c}\exp s(\x_i,\y_j)}.
    \label{eq:app-infonce}
\end{equation}
The expected contrastive objective lower-bounds the data MI \citep{oord2019representationlearningcontrastivepredictive}.

Each block is encoded by a convolutional network.
Its input contains standardized crop values, with variables
outside the block set to zero, together with a block mask.
The encoders produce normalized 256-dimensional embeddings.
Their dot product, multiplied by a learned scale, defines the critic.

We train each critic for 6,000 updates with symmetric InfoNCE,
batch size 2,048, and AdamW.
The maximum learning rate is $10^{-3}$ and weight decay is $10^{-4}$.
A one-cycle schedule uses a 5\% warm-up.
We evaluate every 1,000 updates on $\mathcal D_A$ and select
the checkpoint with the largest bound.

The selected critic is evaluated on disjoint batches from
$\mathcal D_B$ with $N_c\in\{4096,8192,24576\}$.
At $N_c=4096$, we report the mean batch estimate and approximate
one-sided 95\% lower confidence limits, including Bonferroni
correction across the 563 partitions.
We also evaluate a shuffled-$Y$ control.
The InfoNCE objective is capped at $\log_2 N_c$, which equals
12 bits at $N_c=4096$. At $N_c=4096$, the Bonferroni-adjusted one-sided $95\%$ lower confidence limits range from $3.51$ to $11.93$ bits and exceed the learned PC mutual information at every corresponding partition across all tested latent sizes.

\subsection{Model-Selection Protocols}
\label{app:selection-protocols}

\paragraph{HCLT initializations.}
The $h=64$ scatterplot compares 19 initializations on tree $t_1$.
Perturbations of sum and input parameters use
$\lambda\in\{1,4,16,64,128,256\}$, with three seeds at $\lambda=64$.
Perturbations of sum parameters alone use
$\lambda\in\{1,4,16,64,256\}$, with three seeds at $\lambda=16$.
The remaining candidates are three Dirichlet sum initializations
with concentration $0.1$ and one LVD initialization.

We measure the saved checkpoint after two epochs of each
candidate's training run and apply the selection rule
in \cref{sec:model-selection}.
Early validation likelihood is measured on the same samples.
Final likelihood follows the stopping rule in
Appendix~\ref{app:training}.

\paragraph{HMM initializations.}
We use HMMs with $h=512$ on WikiText-103, tokenized with the
GPT-2 tokenizer~\citep{Radford2019LanguageMA} and divided into 128-token sequences.
Full-batch EM uses pseudocount $10^{-6}$.
Each epoch accumulates flows over the training set and applies one parameter update.
The probe is evaluated after two updates on validation
sequences, and likelihood is reported at epoch 200.

The pool contains perturbations of sum and input parameters with
$\lambda\in\{0.1,1,2,4,8,16,32,64,96,128\}$,
two Dirichlet sum initializations with concentrations $0.1$ and $1.0$,
and one LVD initialization.
The score averages over 127 latent groups.
For perturbations through $\lambda=64$, reconstructed early probes
are paired with historical epoch-200 likelihoods from the same
initialization recipe.
For the remaining candidates, the early probe and later likelihood
come from the same initialization.

\paragraph{Tree structures.}
The $h=512$ pool contains ten trees constructed with different
tree-building seeds.
All candidates use $\lambda=64$ uniform-noise perturbation
and the same initialization and training seeds.
We score each tree before training and train every candidate
for 180 epochs.
The final likelihood is the mean of its last three validation
evaluations.

\paragraph{Cross-width comparison.}
The table uses tree $t_0$ as the reference and evaluates
initialization choices on tree $t_1$.
At each width, the initialization pool contains three seeds each
for $\lambda=64$ perturbation of sum and input parameters,
$\lambda=16$ perturbation of sum parameters alone,
and Dirichlet sum initialization.

The current table selects the lowest flow entropy after excluding
candidates whose usage-minus-flow entropy is below $0.75$ times
the pool median.
Final likelihoods are reported only for completed runs.
The selection gap is the likelihood difference between the
selected candidate and the best completed candidate in its pool.

\paragraph{Computing infrastructure.}
We used GPU servers equipped with eight NVIDIA RTX PRO 6000 GPUs per server.

\section{Proofs}
\label{app:proofs}

Information quantities are measured in bits.
The parameter-learning objectives below use natural logarithms.

\subsection{Whole-Model Likelihood Decomposition}
\label{app:whole-model-dependence}

Let $p_{\params}$ denote the full model distribution over $\X=\{X_1,\ldots,X_d\}$.
Recursively partition $\X$ into two disjoint blocks until every block contains one observed variable.
Let $\mathcal H$ denote the collection of splits $(A,B)$ in this hierarchy.
All block distributions below are marginals of $p_{\params}$.

\begin{lemma}[Whole-model likelihood decomposition]
\label{lem:decomp}
For any partition hierarchy $\mathcal H$ of $\mathbf{X}$,
\begin{equation}
\mathbb{E}_{p_{\boldsymbol\theta}}
\left[\log_2\frac{p_{\boldsymbol\theta}(X_1,\dots,X_d)}
                 {\prod_{j=1}^d p_{\boldsymbol\theta}(X_j)}\right]
=\sum_{(\mathbf A,\mathbf B)\in\mathcal H}
 I_{p_{\boldsymbol\theta}}(\mathbf A;\mathbf B).
\label{eq:decomp-lemma}
\end{equation}
\end{lemma}

\begin{proof}
For a split of a block $S$ into disjoint blocks $A$ and $B$, we have
\begin{equation}
\begin{aligned}
    \log_2 p_{\params}(\x_S)
    ={}&
    \log_2 p_{\params}(\x_A)
    +\log_2 p_{\params}(\x_B)\\
    &+
    \log_2
    \frac{p_{\params}(\x_A,\x_B)}
         {p_{\params}(\x_A)p_{\params}(\x_B)}.
\end{aligned}
\label{eq:app-single-split}
\end{equation}
Here, $\x_S$, $\x_A$, and $\x_B$ denote the restrictions of the full assignment $\x$ to the corresponding variable blocks.
Apply this identity recursively, starting from $S=\X$.
Every intermediate block probability cancels, leaving the full joint probability and the univariate marginals.
Consequently, for $p_\theta$-almost every assignment $\x$,
\begin{equation}
\begin{aligned}
    \log_2
    \frac{p_{\params}(\x)}
         {\prod_{j=1}^{d}p_{\params}(x_j)}
    =
    \sum_{(A,B)\in\mathcal H}
    \log_2
    \frac{p_{\params}(\x_A,\x_B)}
         {p_{\params}(\x_A)p_{\params}(\x_B)}.
\end{aligned}
\end{equation}
Taking expectations under $p_{\params}$ gives
\begin{equation}
\begin{aligned}
    \mathbb E_{p_{\params}}
    \left[
        \log_2
        \frac{p_{\params}(X_1,\ldots,X_d)}
             {\prod_{j=1}^{d}p_{\params}(X_j)}
    \right]
    =
    \sum_{(A,B)\in\mathcal H}
    I_{p_{\params}}(A;B).
\end{aligned}
\label{eq:app-whole-model-dependence}
\end{equation}
The expectation of each log-ratio depends only on the joint marginal of its two blocks, which gives the corresponding mutual information.
\end{proof}

The left-hand side is the expected log-likelihood advantage of the full model over the independent distribution with the same univariate marginals.
Any partition of all observed variables can be chosen as the first split, and the remaining terms account for dependence within its two blocks.
The sum in \cref{eq:app-whole-model-dependence} requires a recursive partition hierarchy and is not a sum over all subtree-versus-complement partitions measured in our experiments.

\subsection{Proof of Theorem~\ref{thm:latent-symmetry}}
\label{app:latent-symmetry}

We consider the circuit rooted at the sum node $n$ in \cref{eq:latent-mixture}.
Its normalized distribution is
\begin{equation}
    p_n(\x,\y)
    =
    \sum_{z=1}^{h}
    \params_{n,c_z}\,
    p_n(\x\mid z)\,
    p_n(\y\mid z).
    \label{eq:app-symmetric-mixture}
\end{equation}
The parameter dependence is implicit unless an iteration index is needed.

\paragraph{EM-style updates in terms of flows.}
Following~\cite{liu2026rethinking, vergari2021compositional}, write $\phi_{u,v}=\ln\params_{u,v}$ for an edge from a sum node $u$ to its child $v$.
For a batch $\mathcal B$, define
\begin{equation}
\begin{aligned}
    F_{\phi}^{\mathcal B}(u,v)
    &:=
    \frac{1}{|\mathcal B|}
    \sum_{(\x,\y)\in\mathcal B}
    \frac{\partial\ln\widetilde p_{\phi}(\x,\y)}
         {\partial\phi_{u,v}},\\
    F_{\phi}^{\mathcal B}(u)
    &:=
    \sum_{v\in\ch(u)}F_{\phi}^{\mathcal B}(u,v).
\end{aligned}
\label{eq:app-learning-flows}
\end{equation}
Here, $\widetilde p_{\phi}$ is the raw circuit output, and the derivatives treat edge log-weights as independent coordinates before normalization.
These quantities are posterior expected edge and node counts.
Let $\operatorname{TD}_{\phi}(u)$ denote the probability that a generative traversal of the current normalized circuit visits $u$.

Anemone updates a sum-weight row according to
\begin{equation}
    \params'_{u,v}
    =
    \frac{
        \operatorname{TD}_{\phi}(u)\params_{u,v}
        +\eta F_{\phi}^{\mathcal B}(u,v)
    }{
        \operatorname{TD}_{\phi}(u)
        +\eta F_{\phi}^{\mathcal B}(u)
    },
    \qquad \eta\geq 0.
    \label{eq:app-anemone-update}
\end{equation}
This update and the standard EM updates can be expressed through the common row-normalized form
\begin{equation}
    \params'_{u,v}
    =
    \frac{
        \params_{u,v}
        +\eta T_{\phi}^{\mathcal B}(u)
        F_{\phi}^{\mathcal B}(u,v)
    }{
        1+\eta T_{\phi}^{\mathcal B}(u)
        F_{\phi}^{\mathcal B}(u)
    },
    \label{eq:app-general-em-update}
\end{equation}
where $T_{\phi}^{\mathcal B}(u)$ is a nonnegative multiplier shared by the outgoing edges of $u$.
All quantities on the right are evaluated before the update.

For Anemone, choosing $T_{\phi}^{\mathcal B}(u)=1/\operatorname{TD}_{\phi}(u)$ recovers \cref{eq:app-anemone-update}.
For interpolated mini-batch EM, choosing $T_{\phi}^{\mathcal B}(u)=1/F_{\phi}^{\mathcal B}(u)$ gives
\begin{equation}
\begin{aligned}
    \params'_{u,v}
    &=
    (1-\lambda)\params_{u,v}
    +\lambda
    \frac{F_{\phi}^{\mathcal B}(u,v)}
         {F_{\phi}^{\mathcal B}(u)},\\
    \lambda
    &:=
    \frac{\eta}{1+\eta}.
\end{aligned}
\label{eq:app-minibatch-em-update}
\end{equation}
For full-batch EM, set $\mathcal B=\mathcal D$ and $T_{\phi}^{\mathcal D}(u)=1$, then take $\eta\to\infty$ to obtain
\begin{equation}
    \params'_{u,v}
    =
    \frac{F_{\phi}^{\mathcal D}(u,v)}
         {F_{\phi}^{\mathcal D}(u)}.
    \label{eq:app-full-batch-em-update}
\end{equation}
Thus, full-batch EM is a limiting case of \cref{eq:app-general-em-update}, rather than a finite-step instance of the additive rule.
Rows with zero batch flow are left unchanged, and the Anemone update is applied only to reachable nodes.
Flows at zero weights are interpreted by continuous extension.

\paragraph{Full statement of the symmetry result.}
Suppose the components in \cref{eq:app-symmetric-mixture} have identical parameterizations, with corresponding parameters initialized equally across latent states, and suppose $\params^{(0)}_{n,c_z}=1/h$.
All states use the same batch at each iteration, the same update settings, and symmetrically initialized optimizer states.
Any input-parameter updates and normalization operations also act identically on corresponding parameters.
For gradient-based methods, the optimizer update must commute with permutations of the latent-state parameter blocks.
Assume exact arithmetic and that the likelihoods and updates are well-defined.

Under these conditions, full-batch EM, interpolated mini-batch EM, and Anemone preserve the equality of corresponding component parameters and the uniform mixture weights.
The same holds for gradient-based methods satisfying the stated permutation condition, including SGD, momentum, and Adam with identical settings across states.
Consequently, for every iteration $t$ and every sample in the model's support,
\begin{equation}
    p_n^{(t)}(z\mid\x,\y)
    =
    p_n^{(t)}(z)
    =
    \frac1h,
    \qquad
    I_{p_n^{(t)}}(X;Y)=0.
    \label{eq:app-symmetry-conclusion}
\end{equation}

\begin{proof}
We show that one update preserves the symmetry and then apply induction.

\paragraph{Uniform posteriors give equal flows.}
Suppose the symmetry conditions hold at the beginning of an iteration.
Because corresponding component parameters are equal, there are distributions $a$ and $b$ such that
\begin{equation}
    p_n(\x\mid z)=a(\x),
    \qquad
    p_n(\y\mid z)=b(\y)
\end{equation}
for every state $z$.
Bayes' rule therefore gives
\begin{equation}
    p_n(z\mid\x,\y)
    =
    \frac{h^{-1}a(\x)b(\y)}
         {\sum_{j=1}^{h}h^{-1}a(\x)b(\y)}
    =
    \frac1h.
    \label{eq:app-uniform-posterior}
\end{equation}
Since $n$ is the root of the mixture, its outgoing edge flows satisfy
\begin{equation}
    F_{\phi}^{\mathcal B}(n,c_z)=\frac1h,
    \qquad
    F_{\phi}^{\mathcal B}(n)=1,
    \qquad
    \operatorname{TD}_{\phi}(n)=1.
    \label{eq:app-root-symmetry}
\end{equation}

Now consider corresponding edges $(u_z,v_z)$ and $(u_j,v_j)$ inside two component copies.
Each sample enters the two components with the same posterior probability, and the identical component parameters give identical conditional flows within them.
Their batch flows therefore agree.
Their top-down probabilities also agree because the components have equal mixture weights and identical internal parameters.
Hence
\begin{equation}
\begin{aligned}
    \params_{u_z,v_z}
    &=
    \params_{u_j,v_j},\\
    F_{\phi}^{\mathcal B}(u_z,v_z)
    &=
    F_{\phi}^{\mathcal B}(u_j,v_j),\\
    F_{\phi}^{\mathcal B}(u_z)
    &=
    F_{\phi}^{\mathcal B}(u_j),\\
    \operatorname{TD}_{\phi}(u_z)
    &=
    \operatorname{TD}_{\phi}(u_j).
\end{aligned}
\label{eq:app-corresponding-flows}
\end{equation}
These equalities hold for every batch, not only after averaging over the full dataset.

\paragraph{EM-style updates preserve equal parameters.}
At the root, substituting \cref{eq:app-root-symmetry} into the general update gives
\begin{equation}
    \params'_{n,c_z}
    =
    \frac{
        h^{-1}+\eta T_{\phi}^{\mathcal B}(n)h^{-1}
    }{
        1+\eta T_{\phi}^{\mathcal B}(n)
    }
    =
    \frac1h.
\end{equation}
Inside the components, \cref{eq:app-corresponding-flows} implies that corresponding rows have the same multiplier $T_{\phi}^{\mathcal B}$ for each of the three methods.
Their update numerators and denominators are therefore equal, so corresponding parameters remain equal.
The same conclusion holds in the full-batch EM limit.

Categorical input distributions can be represented as sums over fixed indicators, so their parameters satisfy the same argument.
More generally, corresponding input distributions receive identical posterior-weighted sufficient statistics and remain identical whenever the same deterministic fitting rule is applied to them.

\paragraph{Flow momentum preserves these equalities.}
For a common momentum coefficient $\beta$, the moving-average flow update is
\begin{equation}
    M_{u,v}^{(t)}
    =
    \beta M_{u,v}^{(t-1)}
    +(1-\beta)
    F_{\phi^{(t)}}^{\mathcal B_t}(u,v).
\end{equation}
Equal current flows and equal previous momentum states give equal updated momentum states.
A shared bias-correction factor does not change this equality.
Replacing the flows in the EM-style updates with these corrected moving averages therefore preserves symmetry.
At the root, all outgoing momentum flows are equal, so each remains one $h$-th of their row total.

\paragraph{Gradient-based updates preserve the same symmetry.}
To distinguish likelihood gradients from the raw flows, write
\begin{equation}
    p_n(\x,\y)
    =
    \frac{\widetilde p_{\phi}(\x,\y)}
         {\mathcal Z(\phi)},
    \qquad
    \mathcal L_{\mathcal B}(\phi)
    :=
    \frac{1}{|\mathcal B|}
    \sum_{(\x,\y)\in\mathcal B}
    \ln p_n(\x,\y),
\end{equation}
where $\mathcal Z(\phi)$ is the normalizing constant.
At locally normalized parameters, Lemma~1 of \citet{liu2026rethinking} gives
\begin{equation}
\begin{aligned}
    \frac{\partial\mathcal L_{\mathcal B}}
         {\partial\phi_{u,v}}
    &=
    F_{\phi}^{\mathcal B}(u,v)
    -
    \frac{\partial\ln\mathcal Z(\phi)}
         {\partial\phi_{u,v}}\\
    &=
    F_{\phi}^{\mathcal B}(u,v)
    -
    \operatorname{TD}_{\phi}(u)\params_{u,v}.
\end{aligned}
\label{eq:app-normalized-gradient}
\end{equation}
Thus, corresponding parameters have equal gradients because both their data flows and their model expected edge counts agree.
At the root, the gradient is $1/h-1/h=0$ for every mixture log-weight.
A common gradient-ascent step therefore preserves the equality of corresponding parameters.

The argument also applies when intermediate parameters are not locally normalized.
Let $P_{\pi}$ permute the mixture weights and corresponding component parameter blocks according to a permutation $\pi$ of the latent states.
Relabelling latent states does not change the normalized observed distribution, so
\begin{equation}
    \mathcal L_{\mathcal B}(P_{\pi}\phi)
    =
    \mathcal L_{\mathcal B}(\phi).
\end{equation}
Differentiating gives
\begin{equation}
    \nabla\mathcal L_{\mathcal B}(P_{\pi}\phi)
    =
    P_{\pi}\nabla\mathcal L_{\mathcal B}(\phi).
\end{equation}
At a symmetric parameter vector, $P_{\pi}\phi=\phi$, so the gradient is symmetric as well.
SGD applies the same transformation to these equal gradients.
Momentum and Adam also preserve equality because their corresponding auxiliary states start equal and follow identical update rules.
Any normalization operation that commutes with the same state permutations preserves the resulting symmetry.

\paragraph{Induction gives zero cross-block dependence.}
The initialization satisfies the symmetry conditions, and each update preserves them.
Consequently, at every iteration there are distributions $a_t$ and $b_t$ such that
\begin{equation}
\begin{aligned}
    p_n^{(t)}(\x,\y)
    &=
    \sum_{z=1}^{h}
    \frac1h\,a_t(\x)b_t(\y)\\
    &=
    a_t(\x)b_t(\y)\\
    &=
    p_n^{(t)}(\x)p_n^{(t)}(\y).
\end{aligned}
\end{equation}
The posterior remains uniform by \cref{eq:app-uniform-posterior}, and the factorization implies $I_{p_n^{(t)}}(\X;\Y)=0$.
This proves \cref{eq:app-symmetry-conclusion}.
\end{proof}

\begin{remark}[Internal latent variables]
\label{rem:internal}
The argument extends to an internal sum node $n$, provided its connections and parameter sharing with the surrounding circuit respect the same state symmetry.
Here, flows and top-down probabilities are computed in the full circuit.
Under the symmetry assumptions, \cref{eq:app-root-symmetry} becomes
\[
    F_{\phi}^{\mathcal B}(n,c_z)
    =
    \frac{F_{\phi}^{\mathcal B}(n)}{h}.
\]
Substituting into \cref{eq:app-general-em-update} still gives $\params'_{n,c_z}=1/h$.
The corresponding-flow and momentum arguments above also preserve equality of the component parameters.
Neither $F_{\phi}^{\mathcal B}(n)$ nor $\operatorname{TD}_{\phi}(n)$ needs to equal one.

For gradient-based methods, applying \cref{eq:app-normalized-gradient} to the full-circuit log-likelihood $\mathcal L_{\mathcal B}$ gives, at locally normalized parameters,
\[
    \frac{\partial\mathcal L_{\mathcal B}}
         {\partial\phi_{n,c_z}}
    =
    \frac{
        F_{\phi}^{\mathcal B}(n)
        -
        \operatorname{TD}_{\phi}(n)
    }{h}.
\]
These gradients need not vanish, but they are equal across states.
The permutation argument above therefore preserves equal component parameters and uniform normalized mixture weights under the same optimizer assumptions.
The induction argument then establishes \cref{eq:app-symmetry-conclusion} for the normalized subcircuit distribution $p_n$, with $\X$ and $\Y$ partitioning the scope of $n$.
\end{remark}

\subsection{Proof of Proposition~\ref{prop:latent-information-budget}}
\label{app:latent-information-budget}

All quantities in this proof are evaluated under the model distribution in the proposition.
We omit the distribution subscript for readability.

\begin{proof}
The assumption $\X\perp\Y\mid Z$ gives $I(\X;\Y\mid Z)=0$.
Expanding $I(\X;\Y,Z)$ in two orders yields
\begin{equation}
\begin{aligned}
    I(\X;\Y,Z)
    &=
    I(\X;Z)+I(\X;\Y\mid Z)\\
    &=
    I(\X;Z),\\
    I(\X;\Y,Z)
    &=
    I(\X;\Y)+I(\X;Z\mid\Y).
\end{aligned}
\end{equation}
Equating the two expressions gives
\begin{equation}
    I(\X;\Y)
    =
    I(\X;Z)-I(\X;Z\mid\Y).
    \label{eq:app-shared-information}
\end{equation}

Next, the chain rule applied to the information between $(\X,\Y)$ and $Z$ gives
\begin{equation}
\begin{aligned}
    I(\X,\Y;Z)
    &=
    I(\X;Z)+I(\Y;Z\mid\X),\\
    I(\X,\Y;Z)
    &=
    H(Z)-H(Z\mid\X,\Y).
\end{aligned}
\end{equation}
Therefore,
\begin{equation}
    I(\X;Z)
    =
    H(Z)-H(Z\mid\X,\Y)-I(\Y;Z\mid\X).
\end{equation}
Substituting into \cref{eq:app-shared-information} gives the claimed decomposition,
\begin{equation}
\begin{aligned}
    I(\X;\Y)
    ={}&
    H(Z)-H(Z\mid\X,\Y)\\
    &-
    \bigl[I(\X;Z\mid\Y)+I(\Y;Z\mid\X)\bigr].
\end{aligned}
\end{equation}
\end{proof}
\clearpage
\section{Additional Results}
\label{app:additional-results}

\subsection{Full-Image Comparison with Tensorized Circuits}
\label{app:pic-comparison}

We compare our patch-based HCLTs with the width-512 CP architectures
reported by \citet{gala2024scaling} in Table~D.5.
We evaluate all 50,000 official ImageNet32 validation images and report
bits per dimension (bpd), normalized by $32\times32\times3$.
Each image is divided into four fixed $16\times16$ quadrants, which are
scored independently using the same HCLT parameters.
The full-image log-likelihood is the sum of the four patch
log-likelihoods. This model shares parameters across quadrants but does
not model dependence between them.

For this comparison, we reserve 50,000 images from the training set for
probing and stopping, and train on the remaining images using exact
integer YCoCg-R preprocessing.
Repeating tree selection among ten candidates chooses $t_5$.
On this tree, the entropy-difference gate followed by minimum flow
entropy selects a sum-only Unif-pert(16) initialization from nine
candidates after two epochs, keeping the input initialization unchanged.
All three final HCLT runs reach the 500-epoch cap.
The official validation images are used only for final evaluation in
this comparison.\footnote{These images had been evaluated in earlier
experiments, but did not enter selection or stopping for these runs.}

\begin{table}[t]
    \centering
    \caption{\textbf{Full-image likelihood with training-side model selection.}
    Lower bpd is better. Published width-512 results are from
    \citet{gala2024scaling}, Table~D.5, using lossless preprocessing,
    with five-run means and uncertainties reproduced as reported.
    Gains are positive reductions in bpd
    from tree selection, subsequent initialization selection, and both
    together, calculated before rounding.}
    \label{tab:pic-comparison}
    \begingroup
    \small
    \setlength{\tabcolsep}{8pt}
    \renewcommand{\arraystretch}{1.08}
    \begin{tabular}{llr}
        \toprule
        \textbf{Model / configuration} & \textbf{Optimizer}
        & \textbf{Bpd $\downarrow$} \\
        \midrule
        QT-CP-512, PC  & Adam & $5.286\pm0.001$ \\
        QG-CP-512, PC  & Adam & $5.255\pm0.001$ \\
        QT-CP-512, QPC & Adam & $5.196\pm0.007$ \\
        QG-CP-512, QPC & Adam & $5.085\pm0.001$ \\
        \midrule
        \multicolumn{3}{l}{\emph{Shared HCLT patch product, $h=512$}} \\
        \rowcolor{gray!12}
        $t_0$ + Unif-pert(64) & Anemone & $4.6482$ \\
        \rowcolor{gray!12}
        $t_5$ + Unif-pert(64) & Anemone & $4.6413$ \\
        \quad\emph{$\Delta$ tree selection} &
        & \textcolor{posgreen}{$+0.0069$} \\
        \rowcolor{lightgreen}
        $t_5$ + selected initialization & Anemone & $\mathbf{4.6371}$ \\
        \quad\emph{$\Delta$ initialization selection} &
        & \textcolor{posgreen}{$+0.0042$} \\
        \quad\emph{$\Delta$ combined selection} &
        & \textcolor{posgreen}{$+0.0111$} \\
        \bottomrule
    \end{tabular}
    \endgroup
\end{table}

The HCLT baseline already achieves lower bpd than the reported CP
variants before either selection step (\cref{tab:pic-comparison}).
Several aspects of the models and training differ.
Our HCLT uses a data-learned tree over patch variables and Anemone
training, whereas the published circuits use whole-image quad-tree or
quad-graph structures and Adam with a 50-epoch cap.
Both evaluations use lossless color transforms, but the categorical
encodings also differ, with integer offsets in our implementation and
modulo-256 lifting in theirs.
These differences prevent attributing the cross-paper gap to any one
factor. Within our common HCLT training protocol, tree and
initialization selection reduce bpd by $0.0069$ and a further $0.0042$,
respectively, for a combined reduction of $0.0111$.

\subsection{Dependence across Scope Sizes}
\label{app:mi-scope-size}

\begin{figure}[t]
    \centering
    \includegraphics[width=0.5\linewidth]{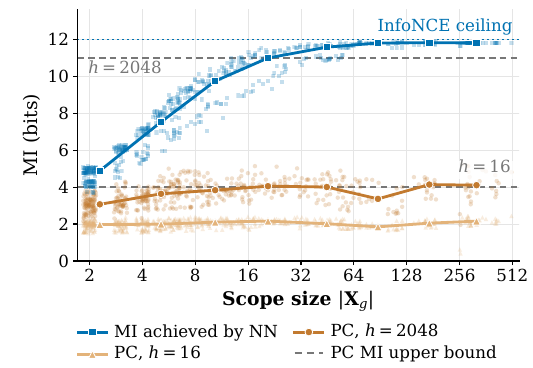}
    \caption{\textbf{The dependence gap is larger for larger scopes.}
    Points compare neural InfoNCE lower-bound estimates with learned HCLT MI across 563 subtree-versus-complement partitions on ImageNet32.
    Points are horizontally jittered, and lines connect medians in logarithmic scope-size bins.
    HCLTs are trained for 180 Anemone epochs.
    Dashed lines mark $\log_2 h$, and the dotted line marks the $12$-bit InfoNCE ceiling ($N_c=4096$).}
    \label{fig:mi-scope-size}
\end{figure}

We group the partitions in \cref{fig:mi-scaling} by scope size $|\X_g|$, the number of observed variables in the subtree below latent $Z_g$, and measure MI between $\X_g$ and its complement $\Y_g$.
The median neural lower-bound estimate rises from $4.89$ bits for scopes of size $2$--$3$ toward the $12$-bit InfoNCE ceiling for larger scopes (\cref{fig:mi-scope-size}).
Across the same size bins, median learned MI varies from $1.85$ to $2.16$ bits at $h=16$ and from $3.07$ to $4.14$ bits at $h=2048$.
The discrepancy is more pronounced for larger scopes, even though learned MI at $h=2048$ remains well below the $11$-bit latent-state upper bound.
The InfoNCE ceiling limits the neural estimate, not the underlying data MI.

\subsection{Dynamics under Other Initializations}
\label{app:initialization-dynamics}

We extend the main-text analysis to LVD and Dirichlet initialization of sum weights, using HCLTs with $h=256$ and the same probing procedure.
These experiments examine which aspects of the observed dynamics persist when training begins from different posterior assignments.

\begin{figure*}[t]
    \centering
    \begin{subfigure}[t]{\linewidth}
        \centering
        \includegraphics[width=\linewidth]{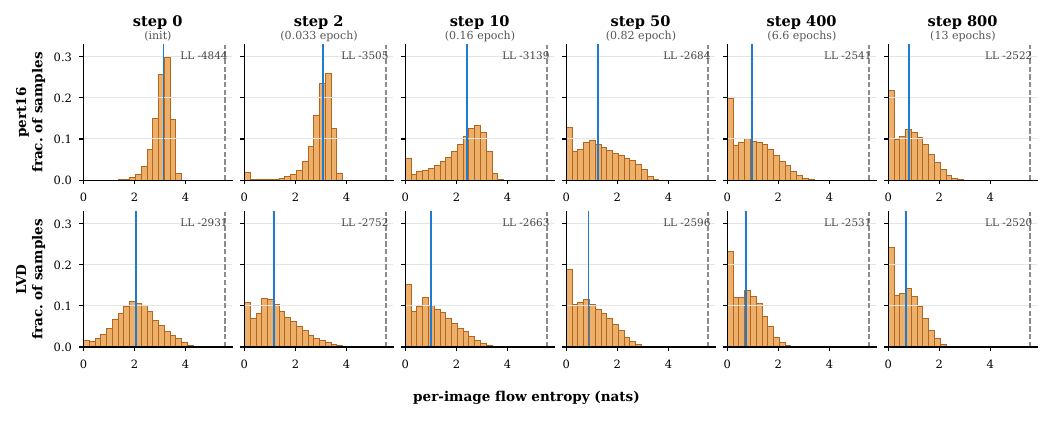}
        \caption{\textbf{LVD begins with more concentrated posteriors.}
        Histograms show per-sample flow entropy for the same latent variable under the two initializations.}
        \label{fig:app-lvd-flow}
    \end{subfigure}

    \vspace{0.4em}
    \begin{subfigure}[t]{\linewidth}
        \centering
        \includegraphics[width=\linewidth]{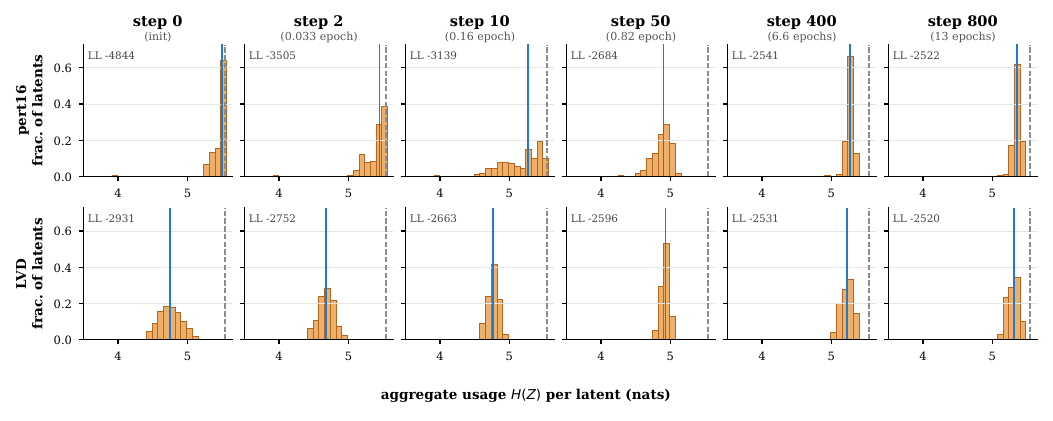}
        \caption{\textbf{LVD begins with less balanced state usage, which broadens during training.}
        Histograms show usage entropy across latent variables.}
        \label{fig:app-lvd-usage}
    \end{subfigure}

    \caption{\textbf{Initialization changes the early entropy trajectories.}
    HCLTs with $h=256$ are trained with Anemone.
    In each panel, the upper row uses uniform-noise perturbation with $\lambda=16$, and the lower row uses LVD.
    Blue lines mark medians, dashed lines mark $\ln h$, and annotations report validation log-likelihood in nats per image.}
    \label{fig:app-lvd-dynamics}
\end{figure*}

\paragraph{LVD starts with more concentrated posteriors and less balanced usage.}
As shown in \cref{fig:app-lvd-dynamics}, LVD produces lower initial flow entropy than uniform-noise perturbation with $\lambda=16$.
For the illustrated latent variable, the median per-sample flow entropy decreases from $2.07$ to $1.14$ nats within two updates and reaches $0.68$ nats at step $800$.
Aggregate usage follows a different trajectory from the perturbation initialization.
The median usage entropy across latent variables starts at $4.75$ nats, decreases slightly to $4.68$ at step $2$, and then rises to $5.31$ at step $800$.
Thus, LVD begins with more concentrated posteriors but less balanced aggregate usage, followed by further posterior concentration and broader state participation.
The pronounced early usage-entropy decrease observed under perturbation initialization is therefore not a universal feature of training.

\clearpage

\paragraph{Dirichlet initialization also exhibits gradual stabilization.}
Dirichlet initialization of sum weights already produces concentrated posteriors before training, but their entropy initially increases.
For the illustrated latent variable, median flow entropy rises from $1.26$ to $1.70$ nats over the first two updates before decreasing to $0.84$ nats at step $800$.
The relative selectivity of samples also changes substantially early in training.
Its rank correlation is $0.33$ between steps $0$ and $2$, compared with $0.66$ between steps $400$ and $800$.
Over the latter interval, $46\%$ of samples have the same dominant state at both checkpoints, increasing to $87\%$ among the most selective fifth.
Low initial entropy therefore does not imply stable state preferences, but the later association between selectivity and persistence also appears under this initialization.

\begin{figure*}[t]
    \centering
    \begin{subfigure}[t]{\linewidth}
        \centering
        \includegraphics[width=\linewidth]{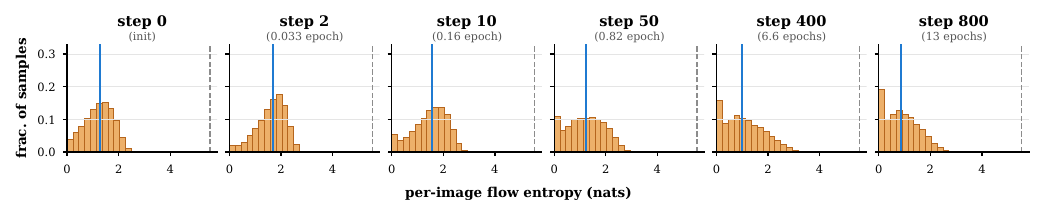}
        \caption{\textbf{Flow entropy initially increases before decreasing.}
        Histograms show per-sample posterior entropy for one latent variable.}
        \label{fig:app-dirichlet-entropy}
    \end{subfigure}

    \vspace{0.4em}
    \begin{subfigure}[t]{\linewidth}
        \centering
        \includegraphics[width=\linewidth]{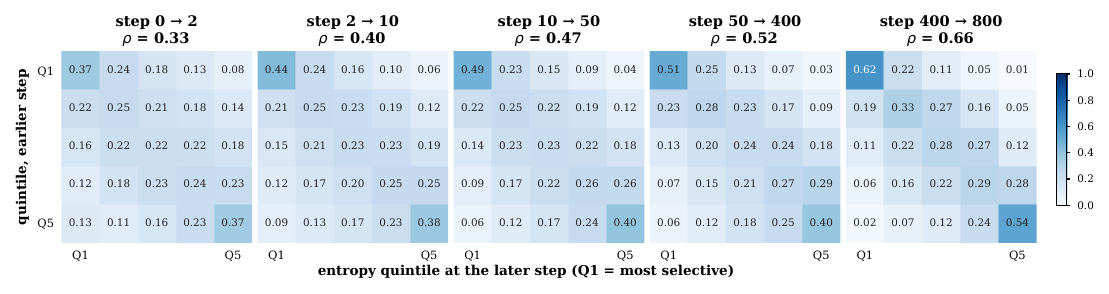}
        \caption{\textbf{The selectivity ranking becomes more stable later in training.}
        Rows and columns indicate earlier and later entropy quintiles, and $\rho$ is the Spearman correlation of per-sample entropies.}
        \label{fig:app-dirichlet-transition}
    \end{subfigure}

    \vspace{0.4em}
    \begin{subfigure}[t]{\linewidth}
        \centering
        \includegraphics[width=\linewidth]{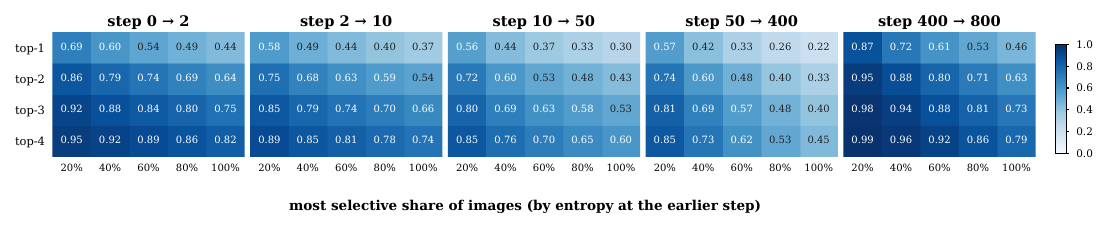}
        \caption{\textbf{Later state preferences are most persistent for selective samples.}
        Columns progressively include more samples in order of increasing earlier flow entropy.
        Entries give the fraction whose earlier dominant state appears among the later top-$k$ states.}
        \label{fig:app-dirichlet-persistence}
    \end{subfigure}

    \caption{\textbf{Posterior reorganization and later stabilization under Dirichlet initialization.}
    Results are for an HCLT with $h=256$ trained with Anemone using Dirichlet-initialized sum weights.}
    \label{fig:app-dirichlet-dynamics}
\end{figure*}

\begin{figure*}[t]
    \centering
    \includegraphics[width=\linewidth]{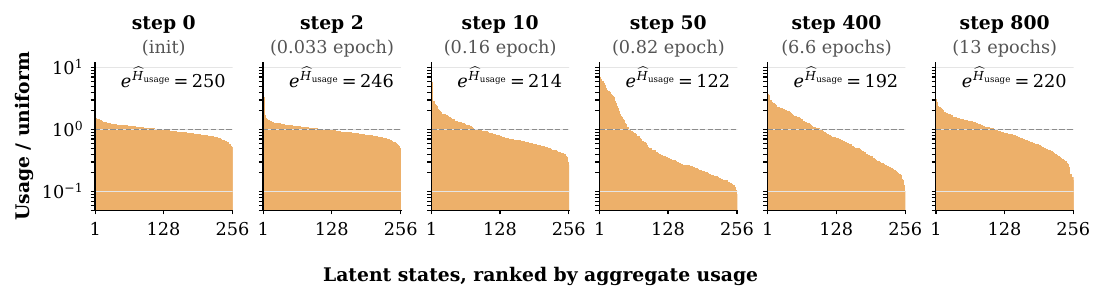}
    \caption{\textbf{Aggregate usage first concentrates and then spreads across more states.}
    The latent variable is the one used in the main-text flow-entropy histograms, with $h=256$ and perturbation strength $\lambda=16$.
    States are sorted independently at each checkpoint by their average posterior probability on the probe set.
    The vertical axis shows $hF_{\phi}^{\mathcal D}(n_z)$, so the dashed line at one denotes equal usage.
    Annotations report the effective number of states, $\exp(\widehat H_{\mathrm{usage}})$.}
    \label{fig:app-usage-profile}
\end{figure*}

\paragraph{State-level usage explains the aggregate entropy trajectory.}
The usage profile in \cref{fig:app-usage-profile} shows how the latent variable used in the main-text flow-entropy analysis distributes posterior mass across its states.
Its effective number of states decreases from approximately $250$ at initialization to $122$ at step $50$, then increases to $220$ at step $800$.
This makes the initial concentration and subsequent broadening of aggregate usage visible at the level of individual states.
States are sorted independently at each checkpoint, so the profiles describe the shape of the usage distribution rather than the identities of the most-used states.

\subsection{Direct Ranking by the Entropy Difference}
\label{app:entropy-difference-selection}

The main-text rule uses $\widehat H_{\mathrm{usage}}-\widehat H_{\mathrm{flow}}$ to filter candidates and then selects the eligible candidate with the lowest flow entropy.
Here, we instead rank candidates directly by this difference and select its maximum.
This comparison tests whether combining the two measurements into a single ranking score changes the selections.

\paragraph{Initialization selection.}
For the 19 HCLT initializations with $h=64$, the epoch-two entropy difference has Spearman correlation $0.66$ with final validation log-likelihood.
Maximizing it selects the same candidate as the main-text rule, with a selection gap of $1.28$ nats per image.
For the 13 HMM initialization settings with $h=512$, the correlation with epoch-$200$ validation log-likelihood is $0.87$, and the selection gap is $0.06$ nats per sequence.
The selected setting again agrees with the main-text rule.
Thus, direct ranking by the entropy difference gives the same choices in these two pools, although it does not perfectly order their final likelihoods.

\begin{figure}[t]
    \centering
    \includegraphics[width=0.5\linewidth]{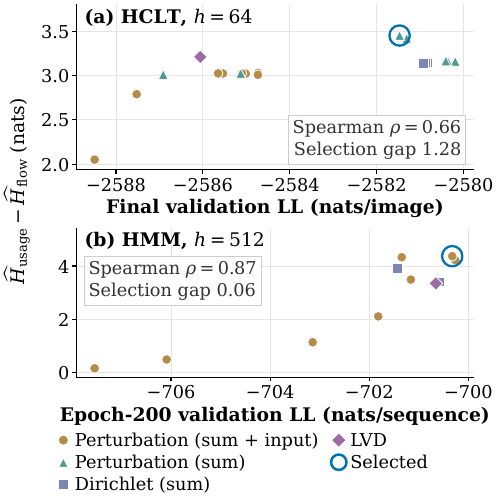}
    \caption{\textbf{Direct entropy-difference ranking gives the same initialization selections as the main-text rule.}
    The difference $\widehat H_{\mathrm{usage}}-\widehat H_{\mathrm{flow}}$ is measured after two epochs.
    Panels show 19 HCLT initializations with $h=64$ and 13 HMM initialization settings with $h=512$.
    Colors and shapes identify initialization families, and blue rings mark the largest difference.
    Boxes report Spearman correlation and the likelihood shortfall from the best candidate in the pool.
    HMM likelihoods are evaluated at epoch $200$, and statistics include all candidates.}
    \label{fig:app-difference-init}
\end{figure}

\begin{figure}[t]
    \centering
    \includegraphics[width=0.5\linewidth]{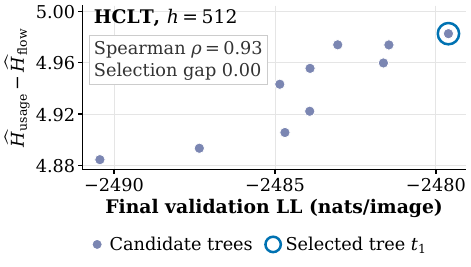}
    \caption{\textbf{The entropy difference and flow entropy give the same tree ranking in this pool.}
    Initial $\widehat H_{\mathrm{usage}}-\widehat H_{\mathrm{flow}}$ is compared with validation log-likelihood for ten HCLT trees with $h=512$ and the same parameter initialization.
    Likelihood is averaged over the last three evaluations of 180-epoch Anemone training.
    The blue ring marks the largest difference, which selects the best-performing tree.
    The box reports Spearman correlation and the selection gap in nats per image.}
    \label{fig:app-difference-tree}
\end{figure}

\paragraph{Tree selection.}
For the ten HCLT trees with $h=512$, the entropy difference measured before training has Spearman correlation $0.93$ with final likelihood and selects the best-performing candidate.
The two scores give identical rankings in this pool, where
usage entropy varies only from $5.992$ to $6.000$ nats.
The agreement in \cref{fig:app-difference-tree} therefore reflects the small variation in aggregate usage within this pool.

\subsection{Comparison with Early Validation Likelihood}
\label{app:early-likelihood}

\begin{figure*}[t]
    \centering
    \includegraphics[width=\linewidth]{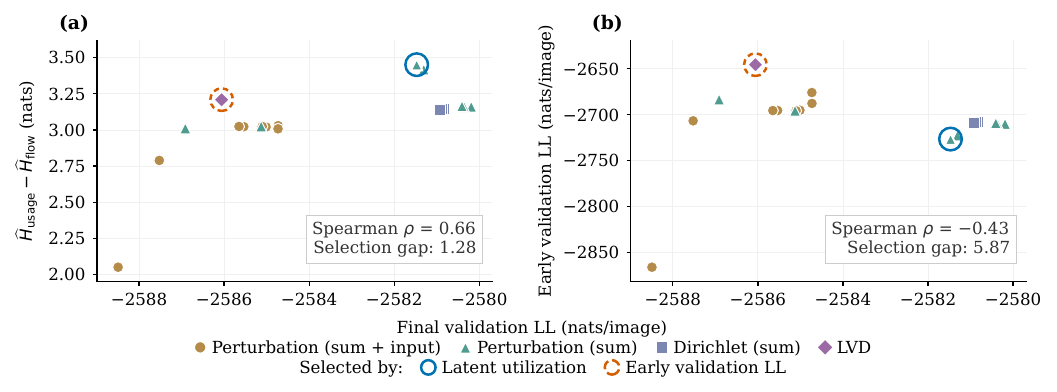}
    \caption{\textbf{Early validation likelihood can misrank initialization strategies.}
    For the same 19 HCLT initializations with $h=64$, panels compare epoch-two entropy difference and validation likelihood with final likelihood after Anemone training.
    Solid blue and dashed orange rings identify the candidates selected by the entropy difference and early likelihood in both panels.
    Boxes report Spearman correlation and the selection gap in nats per image.}
    \label{fig:app-early-likelihood-init}
\end{figure*}

Early validation likelihood evaluates the current model fit, but its ranking need not match the ranking after training.
For the 19 HCLT initializations with $h=64$, validation likelihood after two epochs has Spearman correlation $-0.43$ with final likelihood.
Selecting its maximum chooses the LVD initialization and gives a selection gap of $5.87$ nats per image.
The entropy-based selection instead has a gap of $1.28$ nats per image.
The paired panels in \cref{fig:app-early-likelihood-init} show that the initialization with the highest early likelihood is not the one that achieves the highest likelihood after training.

The additional tree pool with $h=64$ shows that the comparison also depends on probing time.
Before training, flow entropy selects the best of the 15 candidate trees, whereas initial validation likelihood has correlation $-0.52$ with final likelihood and gives a selection gap of $2.12$ nats per image.
After two epochs, validation likelihood in this same tree pool has correlation $0.74$ and selects the best candidate.
These results distinguish initialization selection from tree selection and show why the usefulness of an early likelihood measurement must be assessed at the chosen probing time.

\begin{figure*}[t]
    \centering
    \includegraphics[width=\linewidth]{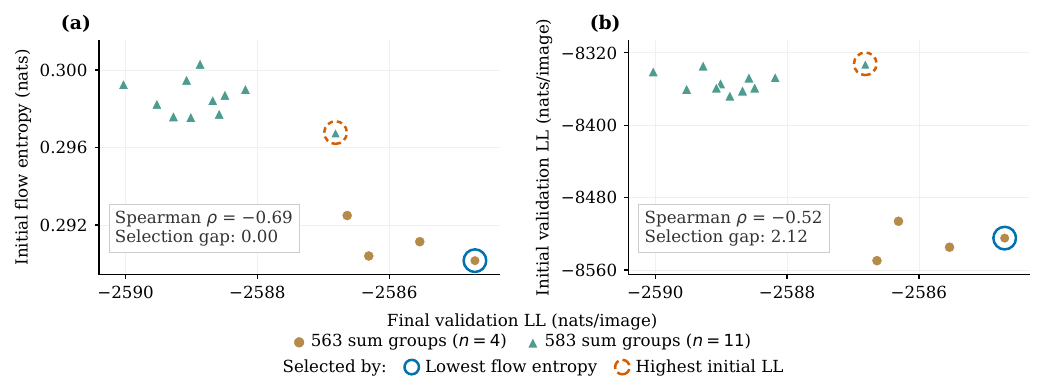}
    \caption{\textbf{Initial flow entropy selects a better tree than initial likelihood in the additional $h=64$ pool.}
    Panels compare flow entropy and validation likelihood before training with final likelihood for 15 candidate trees under the same parameter initialization.
    Colors and shapes distinguish the two sum-node-group counts.
    Rings mark the minimum-flow-entropy and maximum-likelihood selections in both panels.
    Boxes report Spearman correlation and the selection gap in nats per image.}
    \label{fig:app-early-likelihood-tree}
\end{figure*}

\end{document}